\documentclass[authoryear]{elsarticle}
\usepackage{eucal}

\usepackage{hyperref}

\usepackage{mathptmx}       % selects Times Roman as basic font
\usepackage{helvet}         % selects Helvetica as sans-serif font
\usepackage{courier}        % selects Courier as typewriter font
\usepackage{type1cm}        % activate if the above 3 fonts are
\usepackage{dingbat}                           % not available on your system
\usepackage{makeidx}         % allows index generation
\usepackage{graphicx}        % standard LaTeX graphics tool
\usepackage{subcaption}
\graphicspath{{Figures/}}
\usepackage{multicol}        % used for the two-column index
\usepackage[bottom]{footmisc}% places footnotes at page bottom
\usepackage{amsthm}
\usepackage{verbatim}

\theoremstyle{definition}
\newtheorem{definition}{Definition}[section]
\newtheorem{theorem}{Theorem}[section]

\usepackage{filecontents}
\usepackage{booktabs,makecell,multirow}
\usepackage[utf8]{inputenc}
\usepackage[english]{babel}
\usepackage[ruled,longend]{algorithm2e}

\usepackage{xfrac}
\usepackage[group-digits=false]{siunitx}
\usepackage{amsmath}
\usepackage{ amssymb }
\usepackage{subcaption}
\usepackage{geometry}
\usepackage{xifthen}

\makeindex             
\usepackage{tikz}
\usepackage{booktabs}
\usepackage{float}
\floatstyle{plaintop}
\restylefloat{table}
\usetikzlibrary{calc}

\DeclareCaptionLabelFormat{AppendixTables}{A.#2}
\SetKwInOut{Input}{Input}
\SetKwInOut{Require}{Require}
\SetKwInOut{Output}{Output}
\SetKwComment{Comment}{$\triangleright$\ }{}

\date{}

\usepackage{titlesec}

\titleformat{\section}{\normalfont\fontsize{12}{17}\bfseries}{\thesection}{1em}{}
\titleformat{\subsection}
{\normalfont\fontsize{12}{17}\bfseries\itshape}{\thesubsection}{1em}{}

\titleformat{\subsubsection}{\normalfont\fontsize{10}{12}\bfseries}{\thesubsubsection}{1em}{}

\begin{document}
\begin{frontmatter}
\newcommand{\todoSN}[2][]{\textcolor{red}{(SN: #2)\ifthenelse{\isempty{#1}}{}{ #1}}}
\newcommand{\todoOD}[2][]{\textcolor{red}{(OD: #2)\ifthenelse{\isempty{#1}}{}{ #1}}}

%\makecommand*{\email}[1]{%
    %\normalsize\href{mailto:#1}{#1}\par
    %}
    
\clearpage   

\title{No Free Lunch But A Cheaper Supper: A General Framework for Streaming Anomaly Detection}

\author{\normalsize Ece Calikus (Corresponding author)} 
\address{Center for Applied Intelligent Systems Research \\ Halmstad University,Sweden \\ ece.calikus@hh.se}
\author{\normalsize S\l{}awomir Nowaczyk} 
\address{Center for Applied Intelligent Systems Research \\ Halmstad University,Sweden \\ slawomir.nowaczyk@hh.se}
\author{\normalsize Anita Sant'Anna} 
\address{Center for Applied Intelligent Systems Research \\ Halmstad University,Sweden \\ anita.santanna@hh.se}
\author{\normalsize Onur Dikmen} 
\address{Center for Applied Intelligent Systems Research \\ Halmstad University,Sweden \\ onur.dikmen@hh.se}

%\author{Ece Calikus, S\l{}awomir Nowaczyk, Anita Sant'Anna, Onur Dikmen}

%\author{\normalsize Ece Calikus (Corresponding author)} 

%\affil{Center for Applied Intelligent Systems Research \\ Halmstad University,Sweden \\ ece.calikus@hh.se}

%\author{\normalsize Anita Sant'Anna} 
%\affil{Center for Applied Intelligent Systems Research \\ Halmstad University,Sweden \\ \email{user@gmail.com}}

%\author{\normalsize Onur Dikmen} 

%\affil{Center for Applied Intelligent Systems Research \\ Halmstad University,Sweden \\ onur.dikmen@hh.se}

%\institute{Ece Calikus \at Center for Applied Intelligent Systems Research, Halmstad University, Sweden, \email{ece.calikus@hh.se}
%\and S\l{}awomir Nowaczyk  \at Center for Applied Intelligent Systems Research, Halmstad University, Sweden \email{slawomir.nowaczyk@hh.se} \and Anita Sant'Anna  \at Center for Applied Intelligent Systems Research, Halmstad University, Sweden \email{anita.santanna@hh.se} \and Onur Dikmen  \at Center for Applied Intelligent Systems Research, Halmstad University, Sweden \email{onur.dikmen@hh.se}}

%\maketitle

\begin{abstract}
In recent years, research interest in detecting anomalies in temporal streaming data has increased significantly. A variety of algorithms are being developed in the data mining community. They can be broadly divided into two categories, namely general-purpose and ad hoc ones. In most cases, general approaches assume a one-size-fits-all solution model, and strive to design a single ``optimal'' anomaly detector which can detect all anomalies in any domain. To date, there exists no universal method that has been shown to outperform the others across different anomaly types, use cases and datasets. In this paper, we propose SAFARI, a framework created by abstracting and unifying the fundamental tasks within the streaming anomaly detection. SAFARI provides a flexible and extensible anomaly detection procedure to overcome the limitations of one-size-fits-all solutions. Such abstraction helps to facilitate more elaborate algorithm comparisons by allowing us to isolate the effects of shared and unique characteristics of diverse algorithms on the performance. Using the framework, we have identified a research gap that motivated us to propose a novel learning strategy. We implemented twenty different anomaly detectors and conducted an extensive evaluation study, comparing their performances using real-world benchmark datasets with different properties. The results indicate that there is no single superior detector which works perfectly for every case, proving our hypothesis that ``there is no free lunch'' in the streaming anomaly detection world. Finally, we discuss the benefits and drawbacks of each method in-depth,  drawing a set of conclusions and guidelines to guide future users of SAFARI.
\end{abstract}
\end{frontmatter}

\section{Introduction}
\label{sec:1}

Anomaly detection is the problem of identifying data points, or patterns, that do not conform to the expected behavior. Anomalies correspond to (often critical) actionable information in many real-world applications, including condition monitoring, intrusion detection, fault prevention, fraud detection, and so on, across various domains such as production, finance, security, medicine, energy, and social media. In recent years, technological advances have facilitated the ability to collect large volumes of data from streams that are produced by various sensors over time. Therefore, detecting anomalies in such continuously changing temporal data has received increasing attention from both the industry and the scientific community.

However, anomaly detection in data streams is a difficult task, since it combines both the challenges associated with anomaly detection and those associated with learning from streaming data. For example, the former requires defining the exact notion of normal behavior, while the latter includes the difficulty of learning the dynamic nature of such behavior when it evolves over time. Among the many approaches currently proposed in the anomaly detection literature, one can distinguish two categories of methods: general or ad hoc. The ``general'' approaches aspire to detect anomalies independently of the use case and propose a single algorithm supposedly outperforming all previous ones in terms of detection accuracy. However, anomaly detection is an inherently subjective task, in which the characteristics of data and the notion of anomaly vary greatly across applications. One algorithm may perfectly capture the structure of normal behavior in one dataset but may not work at all in another dataset. Several studies show that there is no single anomaly detector that is ultimately superior in all cases \citep{aggarwal2017outlier,campos2016evaluation,emmott2015meta}. Clearly, there is no free lunch for anomaly detection. This fact motivates the need for developing a collection of algorithms instead of seeking for the ``one'' that is suitable all the time.

Ad hoc approaches, on the other hand, are specifically tailored to their target application and are often designed based on complex criteria that require deep domain expertise. Even within the same domain, however, there are often different situations and circumstances where the requirements for a particular task may change. For example, carefully crafted features may become irrelevant, or the current metric to measure deviations in a specific use case may not be suitable in a new scenario. In such cases, making the necessary adaptations to the existing algorithm often amounts to redoing most of the work from scratch.

In this paper, we propose SAFARI, a meta-framework that makes it easy to create different unsupervised anomaly detectors adapted to a particular time-evolving streaming data. This framework provides a generalized procedure for streaming anomaly detection, with separate components that address the fundamental tasks of this problem as separate concerns. The proposed framework is flexible and extensible, since new methods can be easily integrated into existing framework components, to then be mixed and matched for building specific anomaly detectors. Furthermore, the ``loosely coupled'', modularized structure offers a higher degree of freedom for algorithm adaptations, as the properties of each component can be modified separately without the need for updating the other parts of the framework.

Additionally, the existing evaluation strategies do not provide thorough understanding and comparison of proposed algorithms. Most published experiments evaluate their algorithms by reporting performance scores on application-specific case studies or synthetic datasets. They attempt to assess the effectiveness of algorithms without characterizing the nature of the anomalies in the datasets, nor other factors that influence the performance, such as noise or concept drift. Often the methods to be compared share common properties, but it is challenging to analyze their effects on performance because those properties are hidden in the design of the algorithm and are not trivial to isolate. It is not obvious how to interpret whether an algorithm performs better because of, for example, a specific distance function, the sampling strategy, or the features that it uses. This makes it difficult to answer the question \emph{which anomaly detection algorithm should be chosen for a specific scenario?}

By unifying and separating key concepts in existing methods, our framework allows one to study commonalities and differences of the algorithms more thoroughly, leading to more elaborate algorithm comparisons. In this work, we integrate several different methods into the framework and evaluate their performance under varying circumstances. We present how the performances of different combinations vary depending on characteristics of datasets. We also discuss the advantages and drawbacks of each method separately, to guide readers on how to effectively combine building blocks for specific scenarios. In the end, we compare the performances of our framework to those of selected state-of-the-art methods.

Finally, the vast quantity and diversity of existing approaches, as well as difficulty in having an overview of their actual properties, make it challenging to identify the gaps in the state-of-the-art. Our framework helps decomposing many existing approaches in a uniform manner and discovering their essential characteristics. By formalizing state-of-the-art anomaly detection methods within the SAFARI framework, we have determined there is no existing general approach to learning data streams for the case of anomaly detection. Therefore, we propose a novel learning strategy by generalizing the weighted reservoir-sampling schema considering the constraints of the anomaly detection problem.
 
Our contributions can be summarized as follows:
\begin{itemize}
  \item  We conceptualize the four high-level fundamental tasks in streaming anomaly detection problem and formulate a meta-framework that is built upon these essential concepts to provide general, flexible, and adaptable detection procedure.
  %that is built upon
  %\item  We conceptualize the four high-level fundamental tasks in streaming anomaly detection problem and compose these essential concepts into a coherent framework to provide general, flexible, and adaptable detection procedure.
  \item By integrating different methods into the framework’s components, we implement 20 different anomaly detectors, several of which are novel approaches that have not been tried before. 
  \item With the help of the framework, we identify a gap in existing data stream learning strategies and propose a novel anomaly-aware reservoir sampling scheme.
  \item We conduct an extensive comparison study on these approaches using two benchmarks (Numenta and Yahoo) that contain various real-world and synthetic time-series datasets from different domains.
\end{itemize}

The remainder of this paper is organized as follows. In Section 2, we present fundamental concepts in streaming anomaly detection and introduce our framework. We review existing work in Section 3 in the light of concepts introduced in the previous section. In Section 4, we describe in detail the methods we have integrated into the framework implementation. In Sections 5 and 6, we discuss our experiments and results, respectively. We highlight the main observations and recommendations from the results in Section 7 and conclude this study in Section 8.

\section{Framework}
In this section, we present our meta-framework SAFARI (\underline{S}treaming \underline{A}nomaly Detection \underline{F}r\underline{a}mework using \underline{R}eference \underline{I}nstances), which combines fundamental tasks abstracted from existing anomaly detection algorithms into a united schema and provides a generic procedure for streaming anomaly detection. SAFARI is essentially based on the concept of a \textbf{reference group}, which consists a set of instances that is assumed to represent the current normal behavior of the data stream.

\begin{figure}[H]
    \centering
    \includegraphics[width=\textwidth]{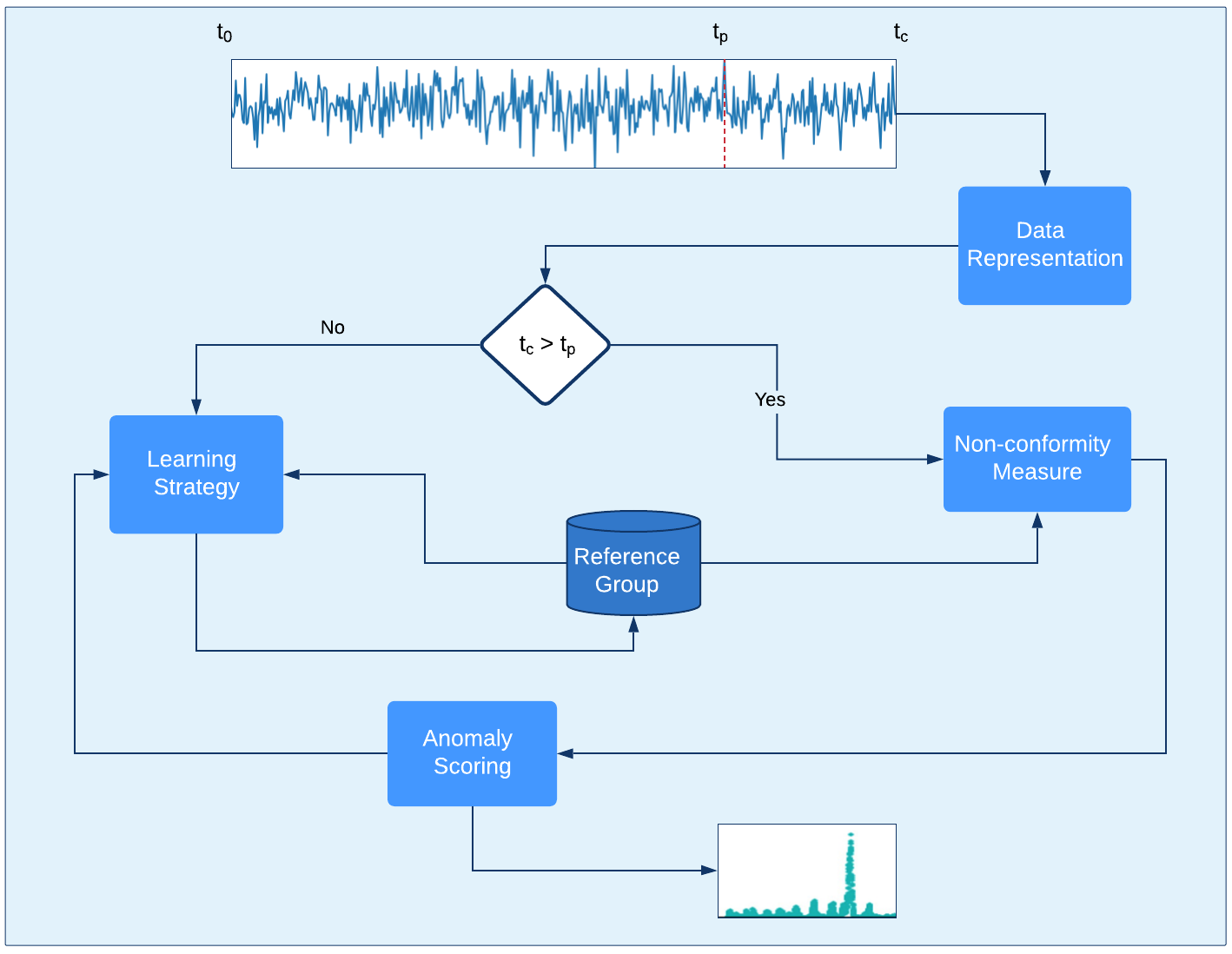}
    \caption{An overview of the SAFARI framework that shows the flow of data and the modeling steps leading to the generation of final anomaly scores. The framework comprises four components along with a decision process where $t_c$ represents the current time and $t_p$  is the time at which the probationary period is ended.}
    \label{framework}
\end{figure}

\newcommand\mycommfont[1]{\footnotesize\ttfamily\textcolor{blue}{#1}}
\SetCommentSty{mycommfont}
\begin{algorithm}
    \SetAlgoLined
    \Input{Stream $S$ = $s_1,s_2,...$ at time $t = 1, 2,...$;\\ Probationary period $p$;\\  Data representations $\mathcal{D}_1,\mathcal{D}_2,... $; \\ Learning strategies $\mathcal{L}_1,\mathcal{L}_2,...$; \\nonconformity measures $\mathcal{A}_1,\mathcal{A}_2,... $;\\Anomaly scorings $ \mathcal{F}_1,\mathcal{F}_2,...$ }
    \Output{Final anomaly scores for every $s_i$ in $S$, where $i>p$}
    $R \leftarrow \{\}$ \Comment*[r]{Reference group}
    $\hat{A} \leftarrow \{\}$ \Comment*[r]{nonconformity scores}
    $\hat{F} \leftarrow \{\}$ \Comment*[r]{Final anomaly scores}
    $\text{Pick a data representation } \mathcal{D} \text{, where } \mathcal{D} \in \mathcal{D}_1,\mathcal{D}_2,... $; \\
    $\text{Pick a learning strategy } \mathcal{L} \text{, where } \mathcal{L} \in \mathcal{L}_1,\mathcal{L}_2,...$; \\
    $\text{Pick a nonconformity measure } \mathcal{A} \text{, where } \mathcal{A} \in  \mathcal{A}_1,\mathcal{A}_2,... $; \\
    $\text{Pick an anomaly scoring } \mathcal{F} \text{, where } \mathcal{F} \in  \mathcal{F}_1,\mathcal{F}_2,... $; \\
     \While{$s_t$ for time $t$ is received}{
        $x_t \leftarrow \mathcal{D}(s_t)$\;
        \eIf{$t \textless p$}{
            $R \leftarrow \mathcal{L}(x_t,R)$ %\Comment*[r]{Compute nonconformity scores of every $x_i \in R$}
            }{
             \eIf(){$\hat{A} = \emptyset$}{
                \For{$x_i \in R$}{
                     $\hat{A} \leftarrow \hat{A} \cup A(x_i,R\setminus x_i)$ \Comment*[h]{Compute the first set of nonconformity scores with the reference group $R$ by leave-one-out fashion}
                }
            }
            {
            $\hat{A} \leftarrow \hat{A} \cup \mathcal{A}(x_t,R)$    \Comment*[r]{Compute the nonconformity score of $x_t$}
            $\hat{F} \leftarrow \hat{F} \cup \mathcal{F}(\hat{A})$ \Comment*[r]{Compute final anomaly score of $x_t$}
            $R_{new} \leftarrow \mathcal{L}_i(x_t,R)$ \Comment*[r]{Update reference group with $x_t$ }
            $\hat{A} \leftarrow \{\}$;\\
            \For{$x_i \in R$}{
                     $\hat{A} \leftarrow \hat{A} \cup A(x_i,R_{new}\setminus x_i)$ \Comment*[r]{Update the nonconformity scores with $R_{new}$}
                }
            
            $R \leftarrow R_{new}$
            }
        }
    }
    \Return $\Phi$ \;
\caption{General Procedure for Streaming Anomaly Detection}
\label{algorithm1}
\end{algorithm}

\subsection{Overview of SAFARI}

SAFARI consists of four main components: data representation (DR), learning strategy (LS), nonconformity measure (NCM) and anomaly scoring (AS), as illustrated in Fig. \ref{framework}. The first component, DR, is concerned with automatically transforming raw input data into informative representations or features, so that it can be effectively exploited in anomaly detection tasks. The second component, LS, deals with the selection of the reference group from transformed data, aiming to extract a representative sample of normal behavior from the stream over time. The third component, NCM, measures the nonconformity score of a single observation, with the goal of quantifying the ``strangeness'' of this observation with respect to the reference group. The last component, AS, aggregates these individual outcomes into a final anomaly score for each observation, taking the global context into account. These four components are explained in more detail in the next section.
  
An overview of the generalized procedure of SAFARI is shown in Alg.~\ref{algorithm1}. It can be seen that the overall procedure is implemented with single-pass constraint—that is, the observations in the data stream are processed one at a time without being stored. Furthermore, we define a fixed probationary period (the time that is required to initialize the framework), with the predictions starting afterwards.

The SAFARI framework is designed based on the ``separation of concerns'' concept, where components are self-contained, cohesive building blocks that serve different purposes in anomaly detection of data streams. This allows one to easily integrate new methods into any of the components or modify the existing components without the need for altering the rest of the framework. Implemented building blocks can be combined in various ways to obtain different and novel detectors, as instantiations of SAFARI.

This setup also provides a basis to conduct more elaborate evaluation experiments. SAFARI allows us to demonstrate, in Section~\ref{sec:eval} of this paper, the contribution of each building block separately and to conduct thorough algorithm comparisons by isolating effects of shared and unique characteristics of different streaming anomaly detection algorithms. Even though some of the existing approaches may not include all four components that we presented here or the reference group approach, most of the streaming anomaly detection algorithms can be unified using SAFARI.

\subsection{The fundamental tasks}
Existing algorithms solve the anomaly detection problem using various approaches and based on different assumptions. However, since the overall goal is to find the instances that do not conform to normal behavior in streaming environments, there are fundamental sub-tasks that are shared among many detectors. We have identified four core concepts that play critical roles in the unsupervised streaming anomaly detection problem and can help in distinguishing the underlying principles of different approaches. Below we present the high-level overviews of these general tasks, and in the next section, we review the state of the art from the perspective of these tasks.

\textbf{Data representation:}
Data representation, in general, is concerned with automatically transforming raw input data into representations or features that can be effectively exploited in machine learning tasks. Useful representations can capture important clues about the past and the current state of the stream as well as the key characteristics of the object (e.g., the monitored system) that are relevant for anomaly detection. The goal of this task is to provide a more vibrant representation of a stream of data consisting of one or more time series that helps to better distinguish anomalies from normal data.

\begin{definition}{(Data representation)}
Let $S=\{s_1,s_2,...,s_t\}$ be an input stream where $s_i \in \mathbb{R}^d$. A \textbf{data representation} is a function $D$ that takes an observation (or a set of observations) in the input stream and transforms them into a feature vector
%\begin{align*}
$x_t = D(s_i,...,s_t)$
%\end{align*}
such that $s_i \in S$, $x_t \in \mathbb{R}^d$. 
\end{definition}
Such features or representations can be obtained in many different ways, such as by extracting means, averages, correlations, or distributions, or by using linear/nonlinear functional relationships, domain knowledge, and so on. There are endless possible model families and hierarchies of models of increasing complexity. 

\textbf{Learning strategy:}

This task is concerned with how to effectively learn the reference group, making sure that it represents the current normal behavior of the stream. A data stream has a continuous flow and the number of incoming observations is unbounded. Unlike static anomaly detection, algorithms that need to learn normal behavior in dynamic environments should have the ability to process new data and limit the number of data points stored in the reference group. At the same time, the reference group should be continuously updated since normal behavior changes over time in a dynamic environment. Therefore, the selection of this set is one of the crucial tasks that differentiate anomaly detection in static and dynamic settings. We refer to the task of maintaining and updating the reference group, where all the observations are not available at once and arrive sequentially, as the ``learning strategy.''  
\begin{definition}{(Learning strategy)} Given that $R_0=\emptyset$, a \textbf{learning strategy} $L$ is a function, 
\begin{align*}
R_t = L(x_t,R_{t-1}),
\end{align*}
where $x_t$ is the current observed feature, $R_{t-1}$ is the reference group before observing $x_t$, and $R_{t}$ is the new reference group at time $t$.
\end{definition}
Various general windowing techniques (e.g., sliding window, damp window) or sampling algorithms (e.g., uniform sampling) can be given as different examples of learning strategies. 

\textbf{Nonconformity measure:}

Essentially, identifying how well the samples conform to the normal behavior is a core step in all unsupervised anomaly detection algorithms. In this task, the goal is to quantify how ``strange'' a single observation is, using a measure which we refer to as ``nonconformity measure''. 

\begin{definition}{(Nonconformity measure)} 
Given the reference group $R_t$ and the sample $x_t$, a \textbf{nonconformity measure} $A$ is a function,
\begin{align*}
a_t=A(x_t,R_t),
\end{align*}
where $a_t$ is the nonconformity score indicating how ``strange'' $x_t$ is with respect to $R_t$. 
\end{definition}
Various approaches with different ``normality'' assumptions can be used to measure nonconformity: for example, measuring the average distances to nearest neighbors, the local density, the variance in the angles, the goodness of fit to a generative model, or the difference between actual and predicted (i.e., expected) values.

\textbf{Anomaly scoring:}
The aim of most anomaly detectors is to output a score for each sample that indicates how likely it is to be an anomaly. In some cases, the levels of ``strangeness'' that are measured in the previous task do not directly correspond to the desired levels of ``anomalousness.'' The candidates with the high nonconformity scores may not be statistically significant or semantically relevant for the particular use-case. This stage is concerned with the post-processing of the nonconformity scores, which transforms ``strangeness'' scores of individual observations into ``anomaly'' scores based on the global context. 

\begin{definition}{(Anomaly scoring)}
Let $A=\{a_1,a_2,...,a_t\}$ be a set of nonconformity scores, an \textbf{anomaly scoring} $F$ is a function 
\begin{align*}
f_t = F(a_i,...,a_t),
\end{align*}
such that $a_i \in A$, that maps nonconformity scores to final anomaly scores.
\end{definition}

For example, in some domains one is interested in collective anomalous behavior of the observations rather than the individual level of ``strangeness.'' The final scoring task can aggregate the nonconformity scores of the samples to produce anomaly scores at the collective level. In another approach, assuming faults or anomalies do not occur suddenly and expecting a certain level of temporal ``continuity'' in the detection, the anomaly scoring task can track nonconformity scores and give higher scores to deviations that persist over time.

\section{Related Work}

To the best of our knowledge, our work is the first attempt to formalize the common tasks and properties of streaming anomaly detection. One similar work was proposed by \citet{schubert2014local}, who discussed similarities and differences in local outlier detection methods, focusing on the notion of ``locality'' and proposed an algorithmic structure that unifies the existing methods. Therefore we review the state of the art related to each of the four core concepts that we introduced in the previous section.

\subsection{Data Representation}

A suitable choice of representation greatly affects the ease and efficiency of all data analysis tasks. Therefore, there is a rich literature around this subject. In this study, we review the techniques that are introduced  primarily for temporal data but that are also suitable to be used in streaming fashion. The most popular techniques widely used for time series representation include the discrete Fourier transform (DFT) \citep{faloutsos1994fast},  piecewise models \citep{geurts2001pattern,yi2000fast}, and singular value decomposition \citep{keogh2001locally}. Many of these techniques have already been extended to be applied in a streaming fashion. For example, \citet{zhu2002statstream} proposed a streaming version of DFT for real-time monitoring of thousands of streams. \citet{lazaridis2003capturing} implemented an online version of piecewise constant approximation with little loss of accuracy. Other methods also aim to summarize data streams using simple statistics (e.g., mean, standard deviation or sum) \citep{cohen2003maintaining}, wavelets \citep{cormode2006fast,gilbert2003one} or histograms \citep{fan2015evaluation}. \citet{bulut2003swat} applied wavelets to represent data streams in a way that is biased towards more recent values.

Linear models are also widely used in many works to represent single or multiple data streams \citep{kargupta2004vedas,d2008diagnostics}.  \citet{kargupta2006board} and \citet{kargupta2010minefleet} used linear correlations to monitor correlations of on-board signals for vehicles. Other studies, by contrast, incorporated methods to capture non-linear relationships such as neural gas models \citep{vachkov2006intelligent} and reservoir computing models \citep{chen2013learning,quevedo2014combining}. Echo state networks were applied by \citet{fan2015evaluation} to represent air compressor sensor data. 
 
There are also more recent techniques that use autoencoders \citep{li2015feature} or neural networks \citep{chen2013learning}. For example, \citet{rognvaldsson2018self} has formalized the ``interestingness'' concept to find useful data representation and include different autoencoders and histograms as representations.   
 
\subsection{Learning strategy}
The large volume of data streams poses unique constraints on the computation process in terms of memory and computations. These challenges have led researchers to propose strategies to efficiently approximate streams over time. While some of these methods are general, in that they can be applied to different data mining tasks (e.g., classification or clustering), the rest are developed to be used in specific problems. The common windowing techniques that are broadly applied for different tasks can be categorized into four groups: landmark, sliding, damped and adaptive.  For example, landmark windows were employed for clustering in Birch \citep{zhang1996birch} and CluStream \citep{aggarwal2003framework} and for the frequent pattern mining \citep{manku_motwani_2012,jin2005algorithm,li2015feature,leung2011frequent}. \citet{subramaniam2006online} and \citet{angiulli2007detecting} used sliding windows for outlier detection, while \citet{yang2009cost} and \citet{rognvaldsson2018self} applied them for condition monitoring. Other methods \citep{cao2006density,leung2011frequent} have incorporated damped windows, in which older data points in the window get lower weights than newer points. Adaptive windows are mostly concerned with change detection where the goal is to adapt to the changes more quickly by changing the size of the window. \citet{bifet2007learning} proposed ADWIN for maintaining a window of variable size, growing automatically when the data is stationary and shrinking when change is taking place.

Different sampling algorithms have also been proposed or adopted for mining data streams. Many streaming outlier detection methods exploit uniform sampling \citep{zimek2013subsampling,liu2012isolation,sugiyama2013rapid}, which is a simple but effective technique. Some others proposed custom sampling techniques that are tailored for specific tasks or algorithms. \citet{kollios2003efficient} proposed a density-biased sampling approach for clustering and outlier detection, in which the probability that a point is included in the sample is determined by the point’s local density. Similarly, \citet{zhang2018density} proposed another density-biased sampling for local outlier detection. While sampling sparser regions at higher sampling rates, it also sampled dense regions, at lower sampling rates, to strengthen ``outlierness'' contrast. 

The effectiveness of most of the general methods—for example, uniform sampling—on anomaly detection have not been well recognized and studied in the literature. On the other hand, it is difficult to identify the benefits of the custom sampling algorithms for anomaly detection, since they cannot be applied outside of the specific settings that they were designed for. For example, the density-biased sampling algorithm of \citet{kollios2003efficient} is only applicable with the methods that use density. We have not come across any learning strategy that is designed to consider the properties of the anomaly detection problem and is also general so that it can be combined with any anomaly detector.

\subsection{Nonconformity measure}
There are many ways to measure nonconformity in anomaly detection problem and there is a huge body of literature on this subject. We suggest several surveys on this topic: \citet{aggarwal2015outlier}, \citet{chandola2009anomaly},\citet{gupta2013outlier} and \cite{zimek2012survey}. Here, we review some common approaches, i.e., probabilistic and statistical proximity-based and prediction-based models that are applied in the streaming setting.

In statistical-based approaches, the aim is to learn a statistical model for a normal behavior of a dataset and determine the nonconformity of new observations by measuring how well they fit into that model. \citet{yamanishi2002unifying} and \citet{yamanishi2004line} proposed SmartSifter, based on an online discounting learning algorithm that incrementally learns the probabilistic mixture model and calculates deviation of the incoming data from this model. Several other statistical methods \citep{kuncheva2011change,song2007statistical} use log-likelihood criteria in order to quantify nonconformity.

The idea in proximity-based methods is to measure nonconformity of data points based on their similarity to or distance from the normal data. \citet{angiulli2007detecting} and \citet{kontaki2011continuous} proposed efficient computation of nearest neighbors and use sliding windows to detect global distance-based outliers in data streams. Distance-based ``local'' outlier techniques that extend the local outlier factor (LOF) algorithm to the case of streaming data have been discussed by \citet{na2018dilof}, \citet{pokrajac2007incremental} and \citet{salehi2016fast}. Many clustering-based methods use distance to the cluster centers as the measure of nonconformity, while proposing varying algorithms to effectively cluster data streams. \citet{cao2006density} used the concept of micro-clusters to distinguish between normal data and outliers based on the distance to the centers. AnyOut \citep{assent2012anyout}, an anytime algorithm, applied a specific tree structure that is suitable for anytime clustering and computes the nonconformity score using the distance to the nearest cluster centroid. \citet{chenaghlou2017efficient} has proposed a hyper-ellipsoidal clustering approach to model the normal behavior of the system, where nonconformity is determined based on the distance to the cluster boundaries.

Prediction-based methods mostly employ regression-based forecasting models, and nonconformity scores are calculated on the basis of deviations between actual observations and their expected (or forecasted) values. Some works used traditional regression methods such as autoregressive modeling, autoregressive moving average, and autoregressive integrated moving average. Since the success of the prediction process greatly affects the final accuracy, most of the prediction-based methods focused on the prediction model rather than the anomaly detection itself. The work in Yahoo's EGADS framework \citep{laptev2015generic} has provided a set of regression methods that can be selected or integrated by the user. Another work by \citet{ahmad2017unsupervised} used hierarchical temporal memory networks as their prediction model and detected anomalies by tracking prediction error over-time. \citet{hundman2018detecting} and \citet{malhotra2015long} have shown that recurrent neural networks achieve high prediction performance and perform effectively across a variety of domains. Several methods have also been focused on speeding up the regression modeling in the context of a large number of data streams and real-time data \citep{huang2007network,jiang2011anomaly,yi2000online}.

\subsection{Anomaly scoring}
Final scoring has not been formalized as a stand-alone process in most of the methods, and therefore it is much more difficult to abstract from the existing approaches. In some works, it appears as a global normalization step—for example \cite{schubert2014local}, which transfers nonconformity scores to anomaly estimates to satisfy a clear gap between the scores of anomalies and normal samples.  \citet{kriegel2011interpreting} and \citet{gao2006converting} proposed generic scoring functions that can convert any set of the nonconformity scores into probability estimates.

\citet{laptev2015generic} discussed that the prediction error (i.e., nonconformity scores) would not be suitable for time-series anomaly detection and computed relative errors in final scoring. An anomaly likelihood function was proposed by \citet{ahmad2017unsupervised} to define how anomalous the current sample is based on the prediction history of the model. A sliding window was maintained on nonconformity scores and the anomaly likelihood of each window was defined as the final anomaly score. \citet{maurus2017let} and \citet{rognvaldsson2018self} applied statistical tests on the nonconformity scores to produce final anomaly scores that capture only deviations that persist over time. A probabilistic approach was proposed by \cite{olsson2015probabilistic}, aggregating point outliers into group (i.e., collective) anomalies. Several other methods used martingales to convert nonconformity measures to change-point estimates \citep{ho2005martingale,ho2010martingale,volkhonskiy2017inductive}.

\section{Methods}
In this section, we present the details of the methods that we have implemented using SAFARI framework \footnote{The code is available at \url{https://github.com/caisr-hh}}. Experimental evaluation of these methods is presented in the next section. In total, we integrate 12 separate methods—namely, two data representations, five learning strategies, four nonconformity measures, and one anomaly scoring method. These are selected so that, using various combinations, we can build 20 different SAFARI anomaly detectors. Even though only one of these methods is new, many of the combinations themselves produce novel anomaly detectors that have not been tried before.

It is important to note that in this study we mainly focus on two out of the four tasks defined in the previous section: learning strategy and nonconformity measure. In most of the existing solutions, the procedures concerning these two essential tasks are entirely embedded in the overall solution, which makes it difficult to study their individual contributions. Furthermore, no studies address the impacts of nonconformity measures and learning strategies separately, nor the impact of combining them into different anomaly detectors. Data representation is heavily investigated in the literature, and we do not believe we can offer important contributions in that area. Furthermore, we have decided to leave the analysis of anomaly scoring to another work, considering that it would greatly increase the number of combinations. Ultimately, we present diverse methods for the tasks of learning strategy and nonconformity measure to investigate their behaviors in depth, while we only implement two data representation methods and one method for anomaly scoring.

\subsection{Data representations}

The first data representation implemented in the SAFARI framework is a simple approach using mean and standard deviation of the last $N$ observations to represent a feature \citep{rodriguez2004support}:

\begin{equation}
\mu_t= \frac{\sum_{i=0}^{N-1} s_{t-i}}{N},
  \end{equation} 
\begin{equation}
\sigma_t= \sqrt{\frac{\sum_{i=0}^{N-1} s_{t-i}-\mu_t}{N}},
  \end{equation} 

where $s_t$ is the observation at time $t$ and $x_t=\{\mu_t,\sigma_t\}$ is the feature at time $t$. 

The second representation method is based on the SAX (symbolic aggregate approximation) \citep{lin2003symbolic} approach: that is, the discretization of the original data stream into symbolic strings. SAX performs this discretization by dividing a z-normalized subsequence into $w$ equal-sized segments. For each segment, it computes a mean value (i.e., piecewise aggregate approximation, PAA, \citealt{keogh2001dimensionality}) and maps it to symbols according to a user-defined set of breakpoints dividing the distribution space into $\alpha$ equi-probable regions, where $\alpha$ is the alphabet size specified by the user.

In this work, we apply SAX on overlapping subsequences in a single-pass streaming fashion. Given a data stream $S=\{s_1,s_2,...,s_t\}$, we generate a SAX word $x_t$, which is the feature at time $t$, based on a subsequence $\hat{s}$ that comprises the last $n$ observations: $\hat{s}=\{s_{t-n-1},s_{t-n}..,s_t\}$.

\subsection{Learning strategies}

Here, we present five different learning strategies that are integrated into the framework. 

The first strategy, fixed reference group (FR), maintains a static set of instances as a reference group—that is, it does not change over time. Clearly, this strategy is not suitable for many streaming scenarios, especially ones with concept drift. However, we include it as a benchmark and to be able to compare static and dynamic methods, showcasing how they perform under different combinations of the framework components.

The other three strategies (i.e., sliding window (SW), landmark window (LW), and uniform reservoir sampling (URES)) are popular techniques that have been widely used in many streaming applications, including classification and clustering tasks. However, a thorough analysis of these approaches in the anomaly detection problem is still missing from the literature. By integrating them, we make it possible to study their individual performances separately from the rest of the framework and identify their benefits or drawbacks in various datasets.

Finally, we propose a new learning strategy, anomaly-aware sampling (ARES), which provides a generic method that requires only anomaly scores as input and is specifically designed considering the research gap in the anomaly detection problem.

\subsubsection{Fixed reference group}
The fixed reference group method maintains a window that collects the observations arriving in probationary period $p$. This learning strategy essentially provides a static reference group that does not change over time after the probationary period is over (Fig. 2a).

\begin{equation}
R_{t} =\begin{cases}
    R_{t-1}, & \text{if $t>p$},\\
    R_{t-1} + \{x_t\}, & \text{otherwise}.
  \end{cases}
 \\
  \end{equation} 

\subsubsection{Landmark window}
In this windowing technique, a fixed timestamp in the data stream is defined as a landmark, and processing is done over data points between the landmark and the present time (Fig. 2c). Landmarks are usually defined by the user where they can be chosen as the starting timestamp of the stream or a specific timestamp such as the beginning of a year. In this study, we assume the landmark is the time when we observe the first sample ($t=0$). 

The reference group at time $t$ with landmark windowing is as follow: 
\begin{equation}
R_{t} =\begin{cases}
    \emptyset, & \text{if $t\leq l$},\\
    R_{t-1}+ \{x_t\}, & \text{otherwise}.
  \end{cases}
 \\
  \end{equation} 
  
where $l$ is the landmark time. 

Note that learning continues by adding the new samples to the reference group unless either the query is explicitly revoked or the stream is exhausted and no additional observations are entered into the system. Therefore, the size of the reference group is not fixed over time.

%\todoSN{I guess somewhere (in problem statement?) you should add that $R_0=\emptyset$}
\subsubsection{Sliding window}
In the sliding window approach, the oldest sample in the window is discarded whenever a new sample is observed (Fig. 2b). Given a window size $w$ and the new observed sample $x_t$, the reference group at time $t$ is updated as below:
\begin{equation}
R_{t} =\begin{cases}
    R_{t-1} - \{x_{t-w}\} + \{x_t\}, & \text{if $t>w$},\\
    R_{t-1}+ \{x_t\}, & \text{otherwise}.
  \end{cases}
 \\
  \end{equation}

\begin{figure}[H]
    \centering
    \includegraphics[width=1.1\textwidth]{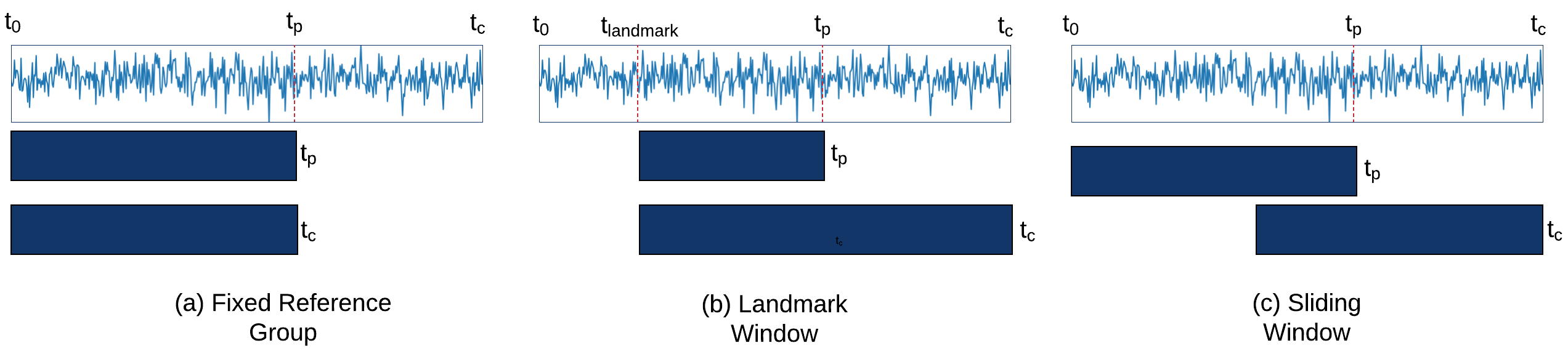}
    \caption{Illustrations of windowing techniques, where $t_c$ represents the current time and $t_p$ is the time where the probationary period is ended.}
    \label{windows}
\end{figure}

\subsubsection{Uniform Reservoir}
The reservoir sampling algorithm \citep{vitter1985random} is a classic method of sampling without replacement from a stream in a single pass when the stream is of indeterminate or unbounded length. Assume that the size of the desired sample is $w$. The algorithm proceeds by retaining the first $w$ items of the stream and then sampling each subsequent element with probability $f(w, t) = \frac{w}{t}$, where $t$ is the current time and also gives the length of the stream so far.

Given reservoir size $w$, the reference group with landmark at time $R_t$ is computed as follows:
 \begin{equation} 
R_{t} =\begin{cases}
    R_{t-1}+  \{x_t \}, & \text{if $t\leq w$},\\
    R_{t-1} -  \{x^* \} +  \{x_t \}, & \text{if $t>w$ $\wedge$ $U < \dfrac{w}{t}$},\\
    R_{t-1}  & \text{otherwise}.\\
  \end{cases}
  \end{equation} 
where $x^*$ is a uniformly chosen element from $R_{t-1}$. 

\subsubsection{Anomaly-aware reservoir}
% The large volume of data streams poses unique space and time constraints on the computation process. 
Most of the learning strategies in the literature focus on providing accurate approximation of the stream while processing large volumes of data efficiently. However, these algorithms, including the other learning strategies integrated into SAFARI, are not designed considering the constraints of the anomaly detection problem and do not guarantee the maintenance of a representative sample of the \emph{normal} behavior over time. For example, the underlying assumption in uniform sampling, which is that all points are of equal importance, has a serious drawback when it is directly applied to a stream containing anomalies. Clearly, sampling the anomalies and normal samples with equal probability can cause the contamination of the reference group and leads to the phenomenon called ``masking,'' which results in the incoming anomalies passing undetected.

To deal with this problem, we propose the anomaly-aware reservoir sampling by generalizing the weighted reservoir sampling schema for anomaly detection problem. In our method, we extend the online algorithm proposed by \citet{efraimidis2006weighted} for the case in which data has a different anomaly score distribution. The goal here is to ensure the samples in the reservoir are more biased toward the samples with lower scores, i.e., samples more likely to be normal.

In a nutshell, the idea of the weighted sampling algorithm is to draw a sample of size $k$ without replacement where the probability of selecting each sample at time $t$ is equal to the sample’s weight divided by the total weights of samples that are not selected before time $t$. Similarly, our learning strategy generates a weighted random sample in one pass over incoming streams and maintains a reservoir with a size $N$ that constitutes the reference group $R$.

The process starts with assigning a ``priority'' to each sample using a weight function $w()$.
Let $x_t$ be the sample at time $t$; we define the function $w(x_t)$ which assigns the weight of $x_t$ as follows:
\begin{equation}
    w(x_t) = exp (-\lambda S(x_t)).
\end{equation}
where $\lambda$ is the decay factor and $S(x_t)$ is the anomaly score of $x_t$. The choices for the decay factor are suggested as $0.96 \leq \lambda \leq 0.98$ by \citet{Haykin:1996:AFT:230061}, and we use $\lambda = 0.96$ in this study. Our method is generic such that $S(x_t)$ can be produced by any method chosen for anomaly scoring, as defined in Section 2.2. Our choice of a method for that task in the experiments is presented in Section 4.4. 

The weight function is designed to give lower importance to instances with high anomaly scores, ensuring that anomalous points have lower probability of being represented in the reference group. Therefore, the strategy aims to avoid learning new abnormal instances while forgetting the ones that are already present. This aspect is especially important when the initial reference group is highly contaminated by the anomalies. 

The learning strategy generates the ``priority'' $p_t = u ^ {\frac{1}{w(x_t)}}$ for the sample $x_t$, where $w(x_t)$ is the weight and $u$ is drawn randomly from [0,1]. In the original implementation, the samples with the highest $N$ priorities are always kept in the reservoir. In each iteration, the sample with the smallest priority is taken as a threshold $T$ and then is replaced by sample $x_t$ if $p_t$ is larger than $T$. 

However, in the presence of a nonstationary distribution, the learning strategy must incorporate some form of forgetting past and outdated information. Therefore, instead of removing the item with the lowest priority, we determine the set of candidate samples that have priorities lower than $x_t$ and remove the oldest one among the candidates. 
 
The goal here is to continuously update the reservoir in such a way that the older items are consistently replaced while still maintaining normal samples in the reference group. 
The details of the overall procedure are shown in Alg. \ref{sampling}

\begin{algorithm}[h]
    \SetAlgoLined
    \Input{Reference group $R$; \\ Reservoir size $w$; \\ New sample $(x_t,t)$}
    \Output{Reference group $R$}
    $priorities \leftarrow \emptyset$\;
    $s_t \leftarrow \text{Collect anomaly score of } x_t$ \;
    $p_t \leftarrow u^{\frac{1}{e^{-\lambda s_t}}}$\;
    \eIf{$t < w$}{ 
        $priorities \leftarrow priorities \cup (p_t, t)$\;
        $R \leftarrow R \cup (x_t, t)$\;
    }{
    $candidates \leftarrow  \text{Collect samples where priorities are smaller than $p_t$}$\;
    \If{$|candidates| > 0$}{
    $i \leftarrow argmin (candidates) $\;
    $priorities \leftarrow priorities / (p_i, i) \cup (p_t, t)  $\;
    $R \leftarrow R / (x_i, i) \cup (x_t, t) $\;
    } 
    }

\Return $R$ \;
\caption{Anomaly-aware reservoir sampling}
\label{sampling}
\end{algorithm}

\subsection{Nonconformity measures}
The four nonconformity measures that are incorporated into the framework are as follows: (i) nearest neighbors-based (NN), (ii) density-based (DEN), (iii) clustering-based (CC) and (iv) frequency-based (FREQ). The first three approaches are based on the popular proximity-based models in which the nonconformity scores are determined by, respectively, the average k-nearest neighbor distance, local density value and distance to closest cluster centroid. Even though these are very common approaches that have been employed by many different anomaly detection algorithms, their streaming versions are not well-studied. 

The fourth method is the frequency-based approach, which measures nonconformity by the number of occurrences of the pattern, with low frequencies leading to higher nonconformity scores. We have integrated this algorithm to increase the diversity in the framework and also to provide a choice that has a significantly lower computational cost.

\subsubsection{Nearest neighbors-based NCM}
The average distances to the k-nearest neighbors (KNN) is used as a measure of nonconformity. 

\begin{equation}
  a_t= \frac{\sum_{i=1}^{k} d(x_t,NN_i(x_t))}{k}, \\
  \end{equation}
where $NN_i(x_t) \in R_t$ is $i$th nearest neighbour of $x_t$.

\subsubsection{Density-based NCM}
This measure quantifies the nonconformity of the samples based on their local densities, under the assumption that anomalies do not lie in dense regions. In this work, we use the LOF to estimate nonconformity scores, since it adjusts for the variations in the local densities of different regions.

Given two points $x_i$ and $x_j \in R$, the $k$-\textit{reachability distance} of $x_i$ with respect to $x_j$ is 
\begin{equation}
R_k(x_i, x_j) = \max\{d(x_i,NN_k(x_i)),d(x_i,x_j)\}, 
  \end{equation}
where $d$ is the distance function and $NN_k(x_i)$ is the $k$th neareast neighbor of $x_i$. 
 \\

Local reachability density, $LRD_k$ is given by  
 \begin{equation}
LRD_k (x_t) = \bigg( \frac{1}{k}\sum_{i=1}^k R_k(x_t,NN_i(x_t)) \bigg) ^{-1},
 \\
  \end{equation}
where $NN_k(x_t) \in R$ is a set of the $k$-nearest neighbors of $x_t$. 

Finally, the nonconformity score of $x_t$ is equal to its local outlier factor, $LOF_k$ given by
\begin{equation}
 a_t = LOF_k (x_t)= \frac{1}{k} \sum_{i=1}^k {\frac{LRD_k(x_t)} {LRD_k(NN_i(x_t))}}.
 \\
\end{equation}

In traditional LOF, the LOF scores of all data points should be updated whenever a new data point is inserted or removed from the reference group $R_t$, which is computationally expensive. We use iLOF \citep{pokrajac2007incremental}, which selectively updates the scores of only the instances affected by the change in the reference group.

\subsubsection{Clustering-based NCM}
In clustering-based NCM, the distance from the nearest cluster centroid is used as a measure of nonconformity. Let $R_t$ be the reference group and $x_t$ be the sample at time $t$, the nonconformity score of $x_t$ is computed as follows:

\begin{equation}
  a_t=min(d(x_t,C(R_t)), \\
  \end{equation}
  where $d$ is the distance function and $C(R_t)$ denotes all the cluster centers computed on $R_t$.

Here, the clustering algorithm used for partitioning the reference group into disjoint sets can be chosen freely. We use the incremental version of k-means by \citet{ordonez2003clustering} to compute clusters and centroids since it can be easily adopted in the streaming scenario. In this method, every new example is added to the cluster with the nearest centroid, and in every $r$ steps a recomputation phase occurs, which updates both the assignment of points to clusters and the centroids. \citet{ordonez2003clustering} chooses $r$ to be the square root of the number of points seen so far, aiming to balance accuracy and computation time. However, in our case, we update the cluster centroids at each time step based on the learning strategy in which the old samples can be removed from the clusters as the new ones are added. Therefore, we follow \citet{bifet2006kalman} for the recomputation phase, who suggests recomputing when an average point distance to centroids has changed more than an $\epsilon$ factor, where the $\epsilon$ factor is user-specified.

\subsubsection{Frequency-based NCM}
This nonconformity measure is motivated by the assumption that anomalies are rare items in the behavior, and samples that form infrequent patterns are more likely to be anomalous. Therefore, measuring nonconformity is directly related to measuring the surprisingness level of the sample, which is defined as the frequency of its occurrence in normal behavior.

After applying the chosen data representation method, the frequency is measured by monitoring the number of occurrences of patterns in the reference group, where a ``pattern'' is a subset of the feature space at any time. Together with this nonconformity measure, we specifically use SAX representation, which has been shown to be a very powerful method to capture meaningful patterns in a data stream \citep{keogh2001locally}. Nonconformity scores of the samples are determined by their ``term'' frequencies—that is, the number of times they occurred in the reference group.

To track term frequencies dynamically, we create a hash table using SAX words encountered in the reference group as the keys and their number of occurrences as hashed values. Given a reference group $R_t$, a hash table $H$ and the current sample $x_t$ that corresponds to a SAX word, the nonconformity score $a_t$ of $x_t$ is computed as

\begin{equation}
a_t=\frac{|R_t|}{f(x_t)+1}. 
\end{equation}
where $|R_t|$ is the size of $R_t$, and $f(x_t)$ retrieves the frequency of $x_t$ from the hash table $H$. The hash table is convenient data structure for this task since insert, update and lookup operations take $O(1)$ and the space is also bounded with $O(N)$ where N is the size of the reference group $R_t$.

\subsection{Anomaly scoring}
In this work, we incorporate only one method for final scoring. It is based on the statistics that has been used in conformal prediction \citep{vovk2005algorithmic}. The procedure of anomaly scoring can be seen in Alg. 3.
To compute anomaly scores, we first estimate p-values for every new observations using nonconformity scores where p-values correspond to confidence levels for each prediction:

\begin{equation}
  p_t= \frac{|i=1,...,w : a_i \geq a_t|}{w}. \\
\label{eq14}\end{equation}
  
In this case, high p-values are consistent with the definition of an outlier by \citet{hawkins1980identification}, where an observation with a high p-value corresponds to the one that deviates so much from other observations as to arouse suspicion that it was generated by a different mechanism. This definition considers an anomaly as an extreme single point that occurs ``individually'' and ``separately.''

In many streaming applications, the temporal continuity plays a critical role to the notion of abnormality, since anomalies mostly occur as abnormal patterns rather than independent outlying observations, or they lead to abrupt or gradual changes exhibiting a lack of continuity with their immediate or long-term history. Furthermore, to be able to detect anomalies in the early stages, one cannot wait for the metric to be clearly beyond the bounds (e.g., p-values) and the ability to detect subtle changes is needed.

We track p-values over time instead of reporting them directly as anomaly scores and apply statistical hypothesis testing under the null hypothesis that the p-values should be uniformly distributed (based on Theorem 1):

\begin{theorem}{\citep{vovk2005algorithmic}}

If the data samples $\{x_1, x_2, \cdots\}$ satisfy the i.i.d. assumption, the p-values $\{p_1, p_2, \cdots \}$  are independent and uniformly distributed in $[0, 1]$.
\end{theorem}

Specifically, this hypothesis is tested using the Kolmogorov-Smirnov (K-S) one-sample test \citep{10030673552}, where we compare the empirical cumulative distribution function of p-values with the cumulative distribution function of the uniform.
 
The empirical cumulative distribution function $F_t(p)$ of the sequence of $n$ p-values $\{p_{t-n+1}, p_{t-n+2}, \cdots ,p_t \}$ is given by  

\begin{equation}
\mbox F_t(p) = \dfrac{1}{n} \sum_{i=t-n+1}^{t} I(p_i \leq p),   \\
\label{eq:devlevel2}
\end{equation}
where $I$ is an indicator function such that $I$ equals 1 if $p_i \leq p$ and 0 otherwise.
%p includes a set of possible values for the random variable $p$
Given $F(p)$ is the cumulative uniform distribution function, the one-sample Kolmogorov–Smirnov statistic for time $t$ is
\begin{equation}
\mbox D_t(p) = sup_p|F_t(p)-F(p)|.  \\
\label{eq:devlevel3}
\end{equation}

where $sup_p$ denotes the supremacy of the set of distances between the curves.

The probability of observing such a $D_t$ under the null hypothesis is evaluated. We use the significance levels obtained from the K-S tests (it should be noted that they are different than the p-values calculated in Eq.~\ref{eq14}) as an indicator for anomaly scores. The significance levels can not be directly interpreted as anomaly scores since p-values will have very low values. Therefore, we apply a score unification step to convert these values into probability estimates by regularization, normalization and scaling steps. Following \cite{kriegel2011interpreting}, we use logarithmic inversion for regularization, a simple linear transformation for normalization and Gaussian scaling to produce final scores. The advantages of the unification of the scores is that it allows the comparison of different combinations of the framework and also makes it possible to create an ensemble of them in the future.

\begin{algorithm}
    \SetAlgoLined
    \Input{Nonconformity scores of the reference group $A_R$; \\ Nonconformity score of the current sample $a_t$; \\ test period $u$}
    \Require{Current p-values $P$;}
    \If(\Comment*[h]{Generate p-values of the first reference group}){$P = \emptyset$}{ 
       \For{$a_i \in A_R$}{ 
        $p_i  \leftarrow  \frac{|j=1,...,|A_R\setminus a_i| : a_j \geq a_i|}{|A_R \setminus a_i|}$\;
        $P \leftarrow P \cup p_i$\;
        }
    }
    $p_t$  $\leftarrow$  $\frac{|j=1,...,|A_R| : a_j \geq a_t|}{A_R}$ \Comment*[r]{Compute p-value of the test sample $x_i$}
    $P \leftarrow P \cup p_t$\;
    $\sigma \leftarrow KSTEST(P,u)$\;
    $s_t \leftarrow UNIFICATION(\sigma)$\;
    
\Return $s_t$ \;
\Output{Anomaly score $s_t$ at time $t$;}
\caption{Anomaly Scoring}
\label{Algorithm3}
\end{algorithm}

\section{Evaluation}
\label{sec:eval}

In this section, we conduct in-depth evaluations for the anomaly detection algorithms within our framework. We introduce the datasets and parameter configurations that we use in this study, and then report our overall evaluation methodology and results. Finally, we summarize our findings and provide intuitive recommendations on selecting appropriate settings for different scenarios.  

\subsection{Datasets}
In the following, we describe the two real-world benchmark datasets—Numenta Anomaly Benchmark (NAB) and Yahoo S5 Webscope Benchmark—that were used in this work.

NAB provides a set of real-world and artificial datasets that are designed for research in streaming anomaly detection. It is composed of 58 datasets containing labeled anomalous periods of behavior. The majority of the NAB datasets are real-world from different domains and applications such as AWS server metrics, Twitter volume, advertisement click metrics, real-time traffic data from Minnesota, temperature sensor data, and so on. Each dataset exhibits different characteristics such as temporal noise, short and long-term periodicities and concept drift.

Yahoo Webscope S5 benchmark is released by Yahoo Labs for the detection of unusual traffic on Yahoo servers. It consists of 367 time-series datasets in four classes in which the ground truth anomaly information is available for all time series. In this study, we use A1 class, which consists of real datasets from Yahoo's computational services, while other classes contain synthetically generated data. A1 datasets comprise 67 time series with various seasonality, distinct change patterns, and diverse types of anomalies that are based on real measurements from various Yahoo cloud services, such as Yahoo Membership Login (YML). 
\subsection{Evaluation metrics}
In our experiments, we adopt two metrics (i.e., ROC-AUC and NAB scoring) to evaluate the detection performances of SAFARI detectors.

The first metric, ROC-AUC, is the most popular measure for the evaluating unsupervised anomaly detection methods. It summarizes the ROC curve score with a single value that ranges between 0 and 1. According to \citet{aggarwal2015outlier} given a scoring of a set of points in order of their propensity to be anomalies, the ROC AUC is equal to the probability that a randomly selected anomaly-nominal pair $(a,n)$ is scored in a correct order where an anomaly appears before a nominal.

\begin{equation} 
    ROC-AUC = mean_{a \epsilon A, n \epsilon N}\begin{cases}
    1, & \text{if $Score(a)>Score(n)$},\\
    1/2, & \text{if $Score(a)=Score(n)$},\\
    0  & \text{if $Score(a)<Score(n)$}.\\
  \end{cases}
\end{equation} 

ROC-AUC is a useful measure to understand whether a method exhibits a high ratio of correctly detected anomalies (i.e., true positive rate, TPR) while providing few normal samples misidentified as anomalies (i.e., false positive rate, FPR). However, this metric only takes the ratio of detected anomalies to nominals into account, ignoring the positions of the samples in the time series.

The second metric that we use in this study is NAB scoring which is a measure provided by Numenta to assess the quality of streaming anomaly detection algorithms. The key aspect of NAB scoring is that it is designed to reward early detection, which is a quite useful feature for many streaming applications. To incorporate the knowledge of early or late detection into scoring, NAB Benchmark defines the concept of an ``anomaly window,'' which consists of a sequence of data points centered on one (or more) true anomalies in a dataset. In a nutshell, NAB scoring considers detection within a window as true positives (TP), which gives positive values to the NAB score such that a TP detected at the beginning of the window has a higher value. If there are multiple detections within a particular anomaly window, the scoring considers only the earliest detection as a TP and ignores all (considered superfluous) detections that follow. This means that an anomaly detector that detects only the first point in the window as an anomaly will receive a higher score than a detector that detects as anomalies all the points in the window except the first one. 

Furthermore, detections made outside the window are considered false positives (FP), and make negative contributions to the NAB score. The position of the detection is also taken into account for FPs. If an FP occurs close to a window, it gets a less negative value than if it occurs further away from the window. Missing a window completely results in a false negative (FN) and makes a strong negative contribution to the score. More details about the method can be found in \citep{lavin2015evaluating}.

The maximum NAB score a detector can achieve in a dataset is equal to the number of anomaly windows in that dataset. To be able to compare detection performances on different datasets, we normalize the NAB scores using the number of windows such that the score of the perfect detector is $1$, and the null detector is $0$. It is important to note that NAB scores are not lower-bounded, since the lowest score of a detector depends on the number of FPs—that is, the number of normal samples in a dataset.

The most important drawback of the NAB scoring is defining anomaly windows efficiently. Selecting larger windows allows the rewarding of earlier detection of anomalies, but it can lead to actual FPs be counted as TPs, thus rewarding inaccurate detection. The authors of the Numenta benchmark \citep{lavin2015evaluating} recommend choosing the window size to be 10\% of the number of instances in a dataset, divided by the number of true anomalies in the given dataset. We follow this suggestion when we generate anomaly windows for each dataset in Yahoo Benchmark to be used for the evaluation with Numenta scoring. 

Contrary to the ROC-AUC score, the NAB scoring requires a threshold value on anomaly scores to cutoff between anomalies and normals. To limit the computational cost, we set a global threshold to $0.9$ providing a guaranteed bound of \%10 false positive rate for SAFARI detectors for all datasets, instead of optimizing the threshold for each dataset separately.

\subsection{Experimental setup}
In our experiments, all requisite parameters of the integrated methods of data representations (i.e., mean-std and SAX), nonconformity measures (i.e., NN, DEN, CC, and FREQ) and anomaly scoring (i.e. CAD) are tuned to select the best parameters for the given evaluation metric. Another parameter of SAFARI, the probationary period, $p$, is chosen as the first 15\% of the total time series for all the datasets as was suggested by the Numenta Benchmark \citep{ahmad2017unsupervised}. Considering this, the window sizes, $w$, required by the learning strategies—FR, SW, URES and ARES—are also set to $w=p$. %The details of how 20 SAFARI detectors that are used in our experiments are built as the combination of 12 SAFARI methods (i.e.,2 DR, 5 LS, 4 NCM and 1 AS) can be found in Appendix A. 

\subsection{Evaluation on Benchmark Datasets}
In this section, we first evaluate the average detection performances of different SAFARI methods, i.e., learning strategies and nonconformity measures across all the datasets. Then, we showcase how the best performances vary among 20 SAFARI detectors (the details are described in Appendix A). %The details on how 20 SAFARI detectors that are used in our experiments are built as the combination of 12 SAFARI methods (i.e.,2 DR, 5 LS, 4 NCM and 1 AS) are presented in Appendix.  

Table \ref{strategy1} presents, for each learning strategy, the NAB and ROC-AUC scores that are aggregated over all datasets combining SAFARI detectors using the same method. More precisely, e.g., the ROC-AUC score of SAFARI-SW indicates the performance of the sliding window method as the mean and the standard deviation of the ROC-AUC results from four different SAFARI detectors that have SW as their learning strategy (i.e., SAFARI-SW-NN, SAFARI-SW-DEN, SAFARI-SW-CC, and SAFARI-SW-FREQ) (see Table \ref{detectors}). The results show that our proposed strategy, SAFARI-ARES, outperforms other methods in both ROC-AUC and NAB scores. SAFARI-FR, as expected, results in the lowest performance.

\begin{table}[H]
\centering
\begin{tabular}{c  @{\hspace*{15mm}}  c@{\hspace*{5mm}} c@{\hspace*{5mm}} c@{\hspace*{5mm}} c@{\hspace*{5mm}} c} 
 \hline
 \\ Performance  & SAFARI-FR & SAFARI-LW &  SAFARI-SW & SAFARI-URES & SAFARI-ARES \\ \\
 \hline
 \\ ROC-AUC & $0.781 \pm 0.15$ & $0.810 \pm 0.13$ &  $0.828 \pm 0.12$ & $0.790 \pm 0.14$ & $\mathbf{0.835 \pm 0.12}$ \\
   \\ NAB  & $0.390 \pm 0.37$ & $0.637 \pm 0.31$ &  $0.629 \pm 0.30$ & $0.559 \pm 0.34$ & $\mathbf{0.660 \pm 0.28}$  \\\\
 Average Rank & $3.71$  & $2.85$  & $2.70$ & $3.16$ & $\mathbf{2.56}$  \\\\
 
\hline
\end{tabular}
\caption{Detection performances of SAFARI’s learning strategies presented using three different metrics: ROC-AUC, NAB and average rank. Results compare the average performances of each method reported as the mean and the standard deviation of the scores taken from all datasets and detectors using that LS. The best average scores across each row of strategies are shown in bold.}
\label{strategy1}
\end{table}

Correspondingly, Table \ref{consensus1} shows the performance comparisons of different nonconformity measures (e.g., SAFARI-NN), averaged over all datasets and all SAFARI detectors with the same NCMs (e.g., SAFARI-FR-NN, SAFARI-SW-NN, SAFARI-LW-NN, SAFARI-URES-NN and SAFARI-ARES-NN). It can be seen that SAFARI-CC achieves the highest performance in ROC-AUC, while SAFARI-FREQ outperforms the others in terms of NAB score. SAFARI-NN consistently leads to the lowest performance.

\begin{table}[H]
\centering
\begin{tabular}{c  @{\hspace*{15mm}}  c@{\hspace*{5mm}} c@{\hspace*{5mm}} c@{\hspace*{5mm}} c} 
 \hline
 \\  Performance & SAFARI-NN & SAFARI-DEN &  SAFARI-CC & SAFARI-FREQ \\ \\
 \hline
 \\ ROC-AUC & $0.767 \pm 0.15$ & $0.820 \pm 0.13$ &  $\mathbf{0.827 \pm 0.13}$ & $0.822 \pm 0.13$  \\
   \\ NAB  & $0.484 \pm 0.36$ & $0.530 \pm 0.34$ &  $0.623 \pm 0.32$ & $\mathbf{0.663 \pm 0.29}$  \\\\
 Average Rank & $3.10$  & $2.53$  & $\mathbf{2.13}$ & $2.22$  \\\\

 \hline
\end{tabular}
\caption{Detection performances of SAFARI’s nonconformity measures presented using three different metrics: ROC-AUC, NAB and average rank. Results compare the average performances of each method reported as the mean and the standard deviation of the scores taken from all datasets and detectors using that NCM. The best average scores across each row of SAFARI-NCMs are shown in bold.}
\label{consensus1}
\end{table}

\begin{figure} 
  \centering
    \includegraphics[width=0.9\textwidth]{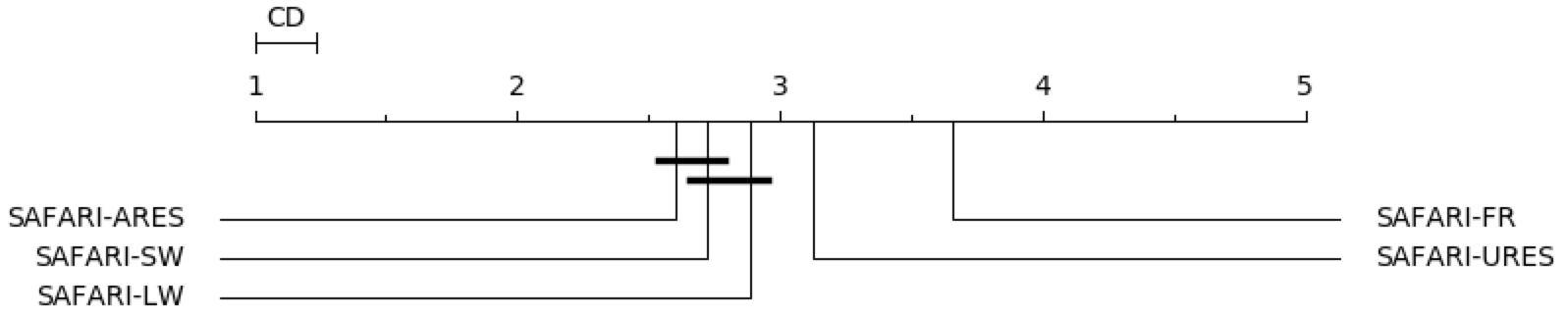}
    \caption{Critical difference diagram showing the streaming anomaly detection performances of the five learning strategies. Methods that are not significantly different (at p < 0.05) are connected with a bar.}
     \label{posthoc1}
\end{figure}

\begin{figure}
  \centering
    \includegraphics[width=0.9\textwidth]{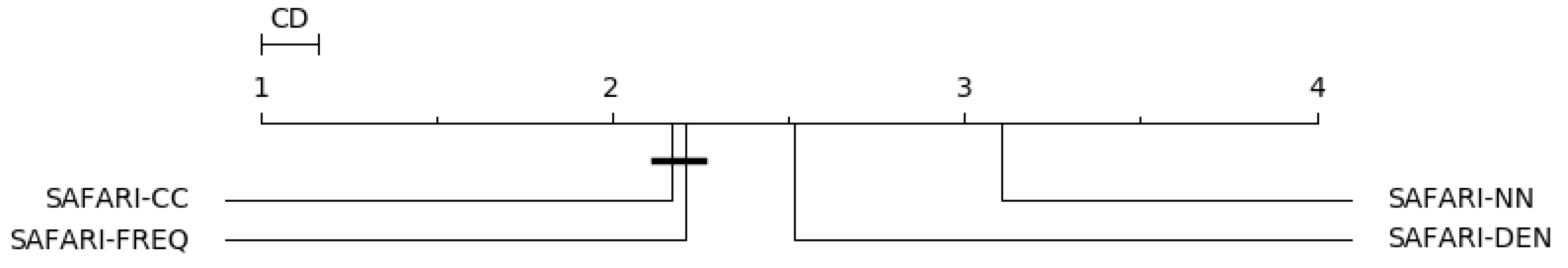}
    \caption{Critical difference diagram showing the streaming anomaly detection performances of the four nonconformity measures. Methods that are not significantly different (at p < 0.05) are connected with a bar.}
    \label{posthoc2}
\end{figure}

To determine whether there is a significant difference between the performances of the different learning strategies and nonconformity measures, we follow \citet{demvsar2006statistical}. We first apply the Friedman test \citep{friedman1937use} using the average ranks of the methods in Table \ref{strategy1} and Table \ref{consensus1} where the null hypothesis for this test assumes that there is no significant difference between the methods. The Friedman tests for learning strategies and nonconformity measures returned p-values of $2.580007E-16$ and $5.758827E-12$, respectively. Therefore, we reject the null hypothesis in both cases and proceed with the Nemenyi post-hoc test \citep{nemenyi1963distribution} to compare methods pairwise and to identify the ones that differ significantly. This test identifies performances of two algorithms to be significantly different if their average ranks differ by at least the ``critical difference'' (CD). Fig. \ref{posthoc1} and \ref{posthoc2} visually represent the results of the Nemenyi tests in
critical difference diagrams where methods that are not connected by a bar have significantly different performances.

For the case of learning strategies, comparing five methods combined with four nonconformity measures on 125 datasets (i.e., 67 Yahoo, 58 Numenta) using two metrics (i.e., ROC-AUC and NAB) at significance level $\alpha = 0.05$, the critical difference diagram is shown in Fig. \ref{posthoc1}. It can be seen that SAFARI-FR performs significantly worse than other learning strategies, demonstrating that a fixed reference group is not a suitable for most of the streaming environments. Furthermore, SAFARI-ARES performs significantly better than SAFARI-FR, SAFARI-LW and SAFARI-URES while the difference between SAFARI-ARES and SAFARI-SW is not statistically significant.

Similarly, in Fig. \ref{posthoc2} we can observe that SAFARI-NN performs significantly worse than the other methods, while there is no significant difference between SAFARI-CC and SAFARI-FREQ. 

In the following, we present how the best performances vary between different SAFARI detectors. Table \ref{combinations} shows, for each combination, the number of datasets for which it gives the best result (in either of the performance metrics). It can be seen that all the combinations achieve the highest performance for at least one dataset, except for SAFARI-FR-NN. Another important observation is that the superiority of a method can be different in terms of average detection performance and the number of best performances. For example, although there is no significant difference among the average performances of SAFARI-CC and SAFARI-FREQ (Fig. \ref{posthoc2}), the number of best performances that SAFARI-FREQ achieves is much higher. In addition, the results show that even the detectors that use the worst methods of the two worlds according to the previous results (i.e., SAFARI-NN as a nonconformity measure or SAFARI-FR as a learning strategy) can achieve the best performances in multiple datasets.

{\tabcolsep=12pt\def\arraystretch{2}}
\begin{table}[H]
\centering
\begin{tabular}{c|cccc|c} 
 \hline
 Combination  & SAFARI-NN & SAFARI-DEN &  SAFARI-CC &SAFARI-FREQ & Total \\ 
 \hline
 SAFARI-FR & $0$ & $4$ & $11$ & $19$ & 34 \\
  SAFARI-LW & $4$ & $11$ & $5$ & $25$ & 45  \\
 SAFARI-SW & $2$  & $14$ & $19$ & $26$ & 60 \\
 SAFARI-URES & $2$ & $7$ & $12$ & $12$ & 33 \\
 SAFARI-ARES & $2$ & $11$ & $17$ & $31$ & $\mathbf{61}$ \\
 \hline
 Total & 10 & 47 & 64 & $\mathbf{113}$ & 234 \\

 \hline
\end{tabular}
\caption{Comparison of the SAFARI detectors based on the number of datasets for which each detector is the winner—that is, outperforms all other detectors. According to results, SAFARI-FREQ-ARES is the detector (combination) with the most wins, with 31 cases. In total, SAFARI-FREQ and SAFARI-ARES are the methods with the highest number of best performances; their results are shown in bold.}
\label{combinations}
\end{table}

The common practice in the anomaly detection literature is comparing different methods based on their ``average'' performances on particular datasets, which is similar to the former results shown in the first part of this section. However, the latter results presented throughout this section show that none of the detectors is able to \textit{consistently} perform better than all the other detectors. This suggests that different combinations are appropriate for different datasets or use cases, even though some of the methods work well more often than others or achieve higher performance on average. In the next sections, we try to highlight which method is likely to be successful under which circumstances.

\subsection{Comparison based on dataset characteristics}

In this section, we discuss and compare the behavior of algorithms across a wide range of datasets with different characteristics. The datasets are delineated based on four properties—noise, concept drift, anomaly type, and anomaly rate. We specifically analyze the individual performances of different nonconformity measures and learning strategies with respect to these properties. The goal is to provide the future users of SAFARI with insights into why combining particular methods may be beneficial or which component is more important for obtaining better results under specific conditions.

We first start by characterizing the datasets and evaluation metrics that are used in this study based on the collective performances of all SAFARI detectors.  For this analysis, we examine the collective performances of all 20 SAFARI detectors and measure their ``difficulty'' and ``diversity'' levels. Following \citet{zimek2012survey}, we define the notion of ``difficulty'' as the average of the scores of all anomalies in the dataset calculated, across all methods. Datasets with a low difficulty score contain anomalies that are relatively easy to detect, while a high difficulty score indicates that the majority of methods have trouble finding the anomalies. On the other hand, ``diversity'' reflects the (lack of) agreement among the detectors on an individual dataset. We define the diversity score of a dataset as the standard deviations of the scores reported by all 20 combinations. A high diversity score indicates a large disagreement among the detection performances.

\begin{figure}[H]
    \centering
    \begin{subfigure}[t]{0.5\textwidth}
        \centering
        \includegraphics[height=2in]{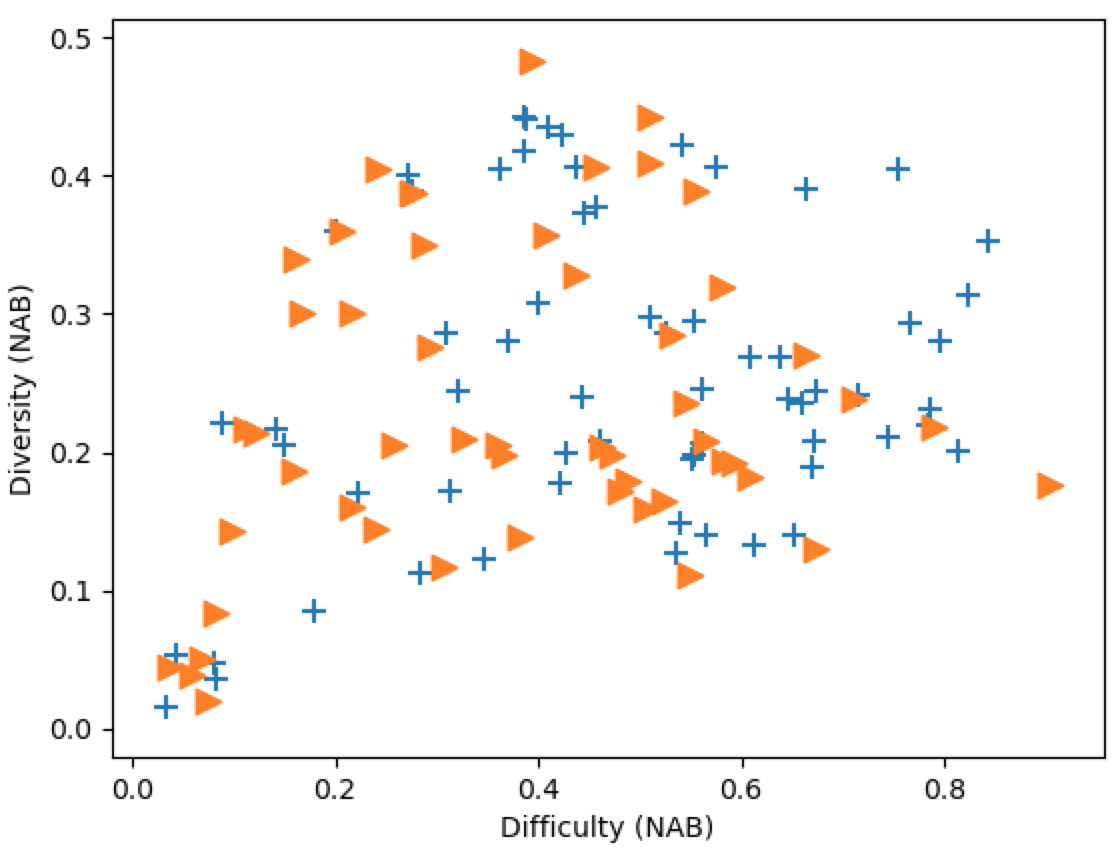}
        \caption{NAB scores}
        \label{dif_nab}
    \end{subfigure}%
    ~ 
    \begin{subfigure}[t]{0.5\textwidth}
        \centering
        \includegraphics[height=2in]{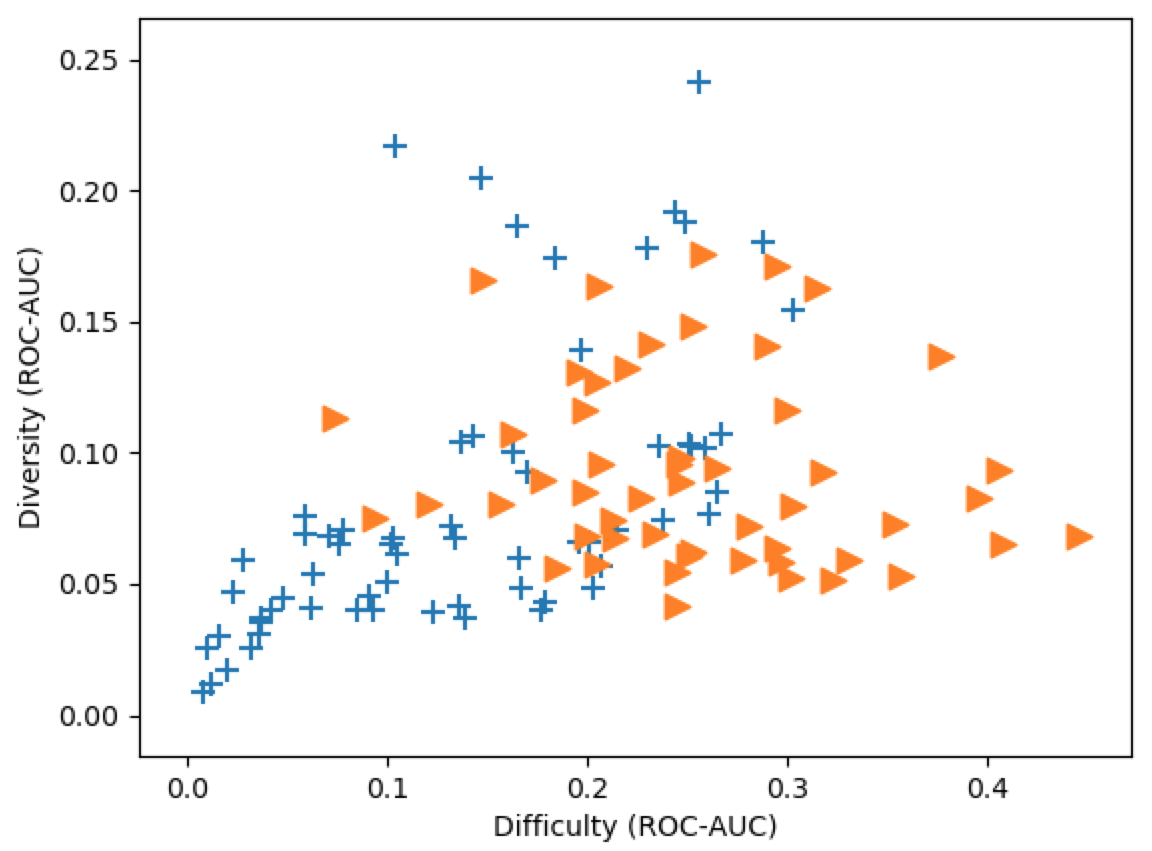}
        \caption{ROC-AUC scores}
        \label{dif_roc}
    \end{subfigure}
    \caption{Diversity versus difficulty of the datasets based on two metrics: NAB and ROC-AUC. Numenta datasets are represented with orange triangles while Yahoo datasets are shown with blue crosses.}
\end{figure}

Figs. \ref{dif_nab} and \ref{dif_roc} show the difficulty–diversity plots using both evaluation metrics. Results from two different benchmarks are represented with different shapes. It can be seen that difficulty and diversity levels can vary greatly between datasets and evaluation metrics. Therefore, making fair comparisons of non-equivalent groups of datasets is not straightforward. For example, suppose we would like to assess the behavior of a method (e.g., sliding window) on a property (e.g., concept drift) by comparing the performance of this method on two groups of datasets: the first group includes ``drifting'' datasets, while the second group includes nondrifting ones. Directly comparing the absolute performances (i.e., the ROC-AUC and NAB score) of the method on these two groups will not be a reliable way to analyze the impact of concept drift, since there can be other factors affecting the performances; in particular, one of the groups is likely to be inherently more difficult. 

In this case, we try to mimic controlled experiments, while our test group (e.g., drifting) and the control group (e.g., non-drifting) have entirely different datasets and therefore, the number of independent variables (factors that are different between two groups) is unknown. To achieve this, we introduce the concept of ``relative performance,'' where the goal is to account for the impact of uncontrollable factors while comparing algorithms performance on a specific property. The relative performance is computed by taking the average difference between the absolute performance of the method and the absolute performances of all the other methods.

It is assumed that the effects of uncontrollable factors also persist in the performance of the other methods, and computing the relative difference between two groups instead of the absolute difference will reduce the effect of this bias. 

More formally, given a dataset $d$, let $S$ be the list of the actual scores of a method $M$, and $\hat{S}$ be the list of actual scores of other methods. The relative performance score of $M$ on $d$ is

\begin{equation}
Rel_d^M = \frac{1}{|S_d|} \sum_{s \in S_d} \frac{\sum_{\hat{s} \in \hat{S_d}} s-\hat{s}}{max(S_d\cup\hat{S_d})-min(S_d\cup\hat{S_d})}.  
 \\
  \end{equation}

Given two sets of datasets $D_1$ (e.g., low-noise datasets) and $D_2$ (e.g., high-noise datasets), the relative performance difference of $M$ between $D_1$ and $D_2$ is 
 
\begin{equation}
\Delta^M= \frac{\sum_{d_i \in D_1} Rel_{d_i}^M}{|D_1|} - \frac{\sum_{d_j \in D_2} Rel_{d_j}^M}{|D_2|}. 
  \end{equation}

Table \ref{major_table} reports relative performance scores of all SAFARI methods (i.e., five LS and four NCM) throughout different dataset properties. Each column represents how a method behaves under certain properties, such as noise, concept drift, and so on. The significant score differences are marked in bold. In the following text we discuss in detail how different properties affect different SAFARI methods.

\textbf{The noise effect:} To compare the effect of noise in the data on the performances of different learning strategies and nonconformity measures, we divide benchmark datasets into two groups: low-noise and high-noise. However, the benchmarks do not provide information regarding the noise level of datasets. Therefore, we have determined this classification through visual analysis of each univariate time series in both Yahoo and Numenta datasets (see supplementary material). 

The relative performance difference ($\Delta$) scores in this setting reflect how the performance of a method changes from high-noise data to low-noise data, in comparison to other methods.

The first column in Table \ref{major_table} shows these scores that are obtained by different SAFARI methods. It can be seen that the impact of noise is not significant in any of the given learning strategies. This result indicates that the choice of the learning strategies is not critical when the level of noise in a dataset is high. On the other hand, the performances of some of the nonconformity measures exhibit significant change under high noise. SAFARI-FREQ has the lowest ($\Delta$) score, which reveals that its performance is the most negatively affected by the increase of noise. On the other hand SAFARI-NN and SAFARI-DEN do not show significant performance decreases between noisy and non-noisy datasets. SAFARI-CC is the most noise resilient method, achieving the highest $\Delta$ score. Considering these finding, SAFARI-CC has clear advantage when there is a clear sign of noise in a dataset, while SAFARI-FREQ should be avoided.

\textbf{Drift effect:} Similar to the previous case, the information about concept drift is missing, therefore we determine it by visual analysis. Following \citep{Gama:2014:SCD:2597757.2523813}, we consider a dataset as drifting qualitatively if it has one of the drift types—that is, sudden, incremental, gradual, or reoccurring. The rest of the datasets are considered as non-drifting (again, see supplementary material). In this setting, a $\Delta$ score indicates how the performance of a method changes from drifting data to non-drifting data in comparison to other methods.

According to Table \ref{major_table}, the drift effect is quite distinct among different learning strategies. $\Delta$ scores show that SAFARI-SW and SAFARI-ARES are better than other methods at dealing with concept drift. Both of these methods have specific forgetting mechanisms, and clearly, forgetting past observations is essential when dealing with drift. The presence of drift affects SAFARI-FR the most, which is expected, considering that it is a static learning strategy that cannot adapt to changes over time.

According to Table \ref{major_table}, most of the nonconformity measures do not show significant performance change between drifting and non-drifting datasets. SAFARI-CC is an exception, exhibiting a clear decrease in performance when datasets are drifting. The explanation of this behavior might be our SAFARI-CC implementation. We use an incremental k-means algorithm that updates clusters over time according to the learning strategy. However, it still assigns a fixed number of clusters ($k$), and if a new concept emerges suddenly, the clustering structure may not adapt well enough to the new concept. This issue can be overcome using a different streaming clustering algorithm to measure nonconformity, one that can also change the number of clusters over time.  

\begin{table}[H]
\centering
\begin{tabular}{c @{\hspace*{20mm}}  c@{\hspace*{5mm}} c@{\hspace*{5mm}} c@{\hspace*{5mm}} c@{\hspace*{5mm}} c} 
 \hline
 DETECTOR  & $\Delta_{noise}$ & $\Delta_{drift}$ &  $\Delta_{type}$ & $\Delta_{contamination}$   \\[5pt] 
 \hline
  SAFARI-FR & $0.0169$ & $\bold{-0.1815}$ & $-0.0081$  & $0.01817$   \\[5pt]
  \hline
 
  SAFARI-LW & $-0.01405$  & $0.0178$  & $0.0194$ & $\bold{-0.0279}$    \\[5pt]
  \hline

  SAFARI-SW & $0.0179$  & $\bold{0.0977}$  & $-0.0189$ & $-0.0184$   \\[5pt]
  \hline

  SAFARI-URES & $-0.0221$  & $0.0279$  & $-0.0070$ & $\bold{-0.0359}$   \\[5pt] \hline
SAFARI-ARES & $0.0012$  & $\bold{0.0362}$  & $0.0176$ & $\bold{0.0538}$   \\[5pt] \hline\\\hline
  
SAFARI-NN & $0.0147$  & $0.0129$  & $0.0001$ & $-0.0138$   \\[5pt] \hline
SAFARI-DEN & $0.0172$  & $0.0380$  & $\bold{-0.0391}$ & $0.0176$   \\[5pt] \hline
SAFARI-CC & $\bold{0.0216}$  & $\bold{-0.0624}$  & $\bold{-0.0237}$ & $0.0181$   \\[5pt] \hline
SAFARI-FREQ & $\bold{-0.0668}$  & $-0.0033$  & $\bold{0.0570}$ & $-0.0219$   \\[5pt]

  \hline

\end{tabular}
\caption{Comparison of the SAFARI methods using relative performance scores across datasets with different characteristics: noise level, concept drift, anomaly type and anomaly rate (contamination). \label{major_table}}
\end{table}

\textbf{Anomaly type effect:} We study the effect of two types of anomalies: clustered (pattern) anomalies and scattered anomalies (outliers). Clustered anomalies mostly occur when the same process generates anomalies multiple times, while scattered anomalies are often generated by different processes. To assess the clusteredness/scatteredness level of anomalies in each dataset, we use the normalized clusteredness measure proposed by \citet{emmott2013systematic}. The normalized clusteredness $nc$ is defined as $log\left(\frac{\sigma_{n}^2}{\sigma_{a}^2}\right)$, where $\sigma_{n}^2$ is the sample variance of the candidate normal points and $\sigma_{a}^2$ is the sample variance of the candidate anomalies. Then, we consider the anomaly type of a dataset as ``scattered'' if $nc \leq 0$ and ``clustered'' if $nc > 0$.

As reported in the third column of Table \ref{major_table}, the performances of the learning strategies do not show any significant difference when the type of the anomaly changes. However, the detection capabilities of different nonconformity measures can be influenced by anomaly type, since they mostly rely on different assumptions of the normality. The results support this argument by showing that most of the nonconformity measures integrated into SAFARI perform significantly differently on scattered and clustered anomalies. For example, the performances of SAFARI-DEN and SAFARI-CC deteriorate significantly when anomalies are clustered. Both of these methods assume that anomalies are located far away from the dense regions, and clustered anomalies can fool these methods by creating dense regions in the space. On the other hand, SAFARI-FREQ is clearly much better than the rest of the methods in handling clustered anomalies because it looks for the occurrence of the ``rare'' patterns rather than outlying individuals.

\textbf{Anomaly rate effect:} Anomaly rate reflects the contamination level of a dataset and is defined by the fraction of observations that are ground-truth anomalies. We divide the datasets into two groups as high and low contamination by considering the average contamination rate in all 112 datasets as a threshold. The datasets with higher rates than the average are categorized as high, while the rest as low contamination.

It can be observed from Table \ref{major_table} that the anomaly rate profoundly affects the behavior of most of the learning strategies. The performances of SAFARI-LW, SAFARI-SW, and SAFARI-URES are significantly worsened when the contamination is high. The likely reason is that these methods learn from data instances without assessing whether they are actually normal observations. The greater the dataset contamination, the more anomalous the behavior these strategies learn. However, our proposed strategy, SAFARI-ARES, is designed to use relative probabilities in order not to learn from potentially anomalous samples. The results show that it is clearly the best method to deal with datasets containing high anomaly rates.

SAFARI-FR also does not seem to be affected by the anomaly rate, which is understandable since it only learns during the probationary periods, which are defined in each dataset to contain only normal instances based on the ground truth. Still, we cannot recommend this strategy because the absolute performance scores of SAFARI-FR are much lower than the rest of the methods in the case of both low and high anomaly rates (see supplementary material).

Finally, no consistent performance change of nonconformity measures is observed between datasets with low and high anomaly rates.

\subsection{Comparison with the baseline algorithms}
In this section, we compare SAFARI with the state-of-the algorithms that are reported by Numenta benchmark. Table \ref{baselines} summarizes the scores of benchmark algorithms across all application profiles (see supplementary material), including the three NAB competition winners \citep{ahmad2017unsupervised}. In addition to the various streaming anomaly detection algorithms, there are three control detectors in NAB. A “null” detector runs through the dataset passively, making no detections, accumulating all false negatives. A “perfect” detector is an oracle that outputs detections that would maximize the NAB score; that is, it outputs only true positives at the beginning of each window. The raw scores from these two detectors are used to scale the score for all other algorithms between 0 and 100. The “random” detector outputs a random anomaly probability for each data instance, which is then thresholded across the dataset for a range of random seeds. The score from this detector offers some intuition for chance-level performance on NAB.

\begin{table}[h!]
\centering
\resizebox{\columnwidth}{!}{
\begin{tabular}[h!]{c @{\hspace*{20mm}}  c@{\hspace*{10mm}} c@{\hspace*{10mm}} c} 
 \hline
 Detector  & Standard Profile & Reward Low FP & Reward Low FN \\ 
 \hline
 Perfect & 100 & 100 & 100 \\ 
 SAFARI-Best & 91.65 & 88.5 & 95.8 \\
 SAFARI-LW-CC  & 71.75 & 69.1 & 77.8 \\
 Numenta HTM & 70.1 & 63.1 & 74.3 \\
 CAD-OSE & 69.9 & 67 & 73.2 \\
 Numenta & 64.6 & 58.8 & 69.6 \\ 
 KNN-CAD & 58.0	& 43.4	& 64.8 \\
 SAFARI-Average  & 55.5 & 49.1 & 60.8 \\
 Relative Entropy & 54.6 & 47.6 & 58.8 \\ 
 HTM PE & 53.6 & 34.2 & 61.9 \\
 Random Cut Forest & 51.7 & 38.4 & 	59.7 \\
 Twitter ADVec & 47.1 & 33.6 & 53.5 \\
 Etsy Skyline & 35.7 & 27.1 & 44.5 \\
 Sliding Threshold & 30.7 & 12.1 & 38.3 \\
 Bayesian Changepoint & 17.7 & 3.2 & 32.2 \\
 EXPoSE & 16.4 & 3.2 & 26.9 \\
 Random & 11 & 1.2 & 19.5 \\
 Null & 0 & 0 & 0 \\
 \hline
\end{tabular}
}
\caption{Comparison of SAFARI with algorithms in NAB scoreboard}
\label{baselines}
\end{table}

SAFARI-Best in Table \ref{baselines} represents the best combination giving the highest NAB score in each dataset while SAFARI-Average reports the average NAB score of all the combinations. We have also reported the best SAFARI detector across all NAB datasets, SAFARI-LW-CC, which combines distance to cluster centroids as a nonconformity measure and landmark window as a learning strategy.
    
Overall we can observe that SAFARI-Best and SAFARI-LW-CC outperform all other algorithms, while SAFARI-Average delivers competitive results. Numenta HTM, CAD-OSE, Numenta and KNN-CAD are the other detectors that perform well 

\section{Main Observations and Recommendations}
According to the above comprehensive evaluations covering different aspects of anomaly detection, we can conclude that each approach has its own merits and weaknesses. In the following, we provide a summary of our findings and recommend for future SAFARI users potential ways to combine building blocks for specific cases.

First of all, SAFARI-ARES and SAFARI-SW as learning strategies and SAFARI-CC and SAFARI-FREQ as nonconformity measures outperform their competitors in terms of average performance across all the datasets. SAFARI-FR is the significantly worst method, which confirms the prior assumption that static learning is not suitable for streaming scenarios. On the other hand, it was unexpected to observe that SAFARI-NN performed significantly worse than the other nonconformity measures, since the nearest neighbor-based methods showed clear advantages in static datasets in the past \citep{aggarwal2017outlier}. It is important to note that our experiments do not reflect the parameter sensitivity of the methods. We recommend users to refer to the studies by \citet{aggarwal2017outlier}, \citet{campos2016evaluation}, and \citet{goldstein2016comparative} if they would like to consider the stability of the algorithms across a wide range of parameter choices.

From the perspective of different dataset properties, we observed that the choice of learning strategy is particularly important if datasets include concept drift or high anomaly rate. These properties can influence the performances of different learning strategies in different manners. While SAFARI-SW is the best method under concept drift, which shows the importance of adapting to the newest behavior, SAFARI-ARES also achieves competitive results. Furthermore, we recommend users choose SAFARI-ARES if the datasets are highly contaminated with abnormal samples or if it is difficult to obtain normal samples to initialize the model.

We have also found that the noise level and anomaly type of datasets have significant impacts on the performances of nonconformity measures, while we did not observe much effect on learning strategies. Specifically, SAFARI-CC is the most noise resilient method, while SAFARI-FREQ performs consistently worst under high noise. Regarding different types of anomalies, we recommend users consider SAFARI-CC and SAFARI-DEN for scattered anomalies and SAFARI-FREQ for anomalies that are more clustered. 

\section{Conclusion}

In this paper, we introduced SAFARI, a framework for streaming anomaly detection based on building-blocks derived from fundamental concepts of this problem. By combining SAFARI’s adaptive and extensible components, we produced 20 different anomaly detectors, a number of which are novel variants that, to the best of our knowledge, have never been tried before. 

We have conducted comprehensive evaluation studies on these detectors using real-world benchmark datasets. We have discussed their merits and drawbacks thoroughly and drawn a set of interesting take-away conclusions. We have discovered that learning strategies should be chosen carefully for the cases where datasets are suspected of having concept drift or a high level of contamination. SAFARI-SW and SAFARI- ARES are safer methods under concept drift, and SAFARI-ARES is the best option for highly contaminated datasets. Similarly, the selection of nonconformity measures is more critical if datasets include noise or different types of anomalies. Based on a detailed performance analysis, SAFARI-CC is recommended when the dataset has a high level of noise and anomalies are scattered, while SAFARI-FREQ is a better option for clustered anomalies.        

The results have shown that there is no single superior detector that works well for every case and have proven our initial hypothesis that ``there is no free lunch'' in the streaming anomaly detection world. Furthermore, we have also showcased how SAFARI could help to ease this problem by empowering us to easily create use-case-specific detectors that are suitable for different scenarios instead of blindly relying on a single method. 

Finally, we have postulated the problem of generalization and abstraction of streaming anomaly detection by considering similarities and differences in existing approaches. We believe that formally identifying core tasks as building blocks will help in understanding existing or new methods from a unified perspective and lead to identifying research gaps and unattended problems. With the help of SAFARI,  we have discovered such a gap and formulated a new learning strategy specifically designed to handle high contamination while learning the normal group.  

\section*{Acknowledgements}
This research is supported by the Swedish Knowledge Foundation (KK-stiftelsen) \big[Grant No. 20160103\big].

\appendix
\section*{Appendix A}
\setcounter{table}{0}
\renewcommand{\thetable}{A\arabic{table}}
The details of how 20 SAFARI detectors that are used in our experiments are built as the combination of 12 SAFARI methods (i.e., 2 DR, 5 LS, 4 NCM, and 1 AS) can be found in Table \ref{detectors}. In this study, we mainly focus on learning strategy and nonconformity measure, as explained in Section 4. Therefore, we integrate different methods into the components of SAFARI dealing with these two tasks, while we implement only two data representation methods and one method for anomaly scoring. Furthermore, each data representation is combined with specific nonconformity measures in order to limit the number of combinations as 20. The first data representation, mean-std, is used where nonconformity measure is one of the proximity-based methods—nearest neighbors-based, density-based, and clustering-based, while SAX is found more suitable for frequency-based NCM considering its usefulness in capturing subpatterns in time-series \cite{keogh2001locally}.

\begin{table}[H]
\centering
\resizebox{\columnwidth}{!}{
\begin{tabular}{ccccc} 
 \hline
 Detectors & Data Representation & Learning Strategy & Non-conformity Measure & Anomaly Scoring \\ 
\hline
 SAFARI-FR-NN & Mean-Std & Fixed reference group &  Nearest neighbors-based & Final scoring  
  \\ SAFARI-FR-DEN  & Mean-Std & Fixed reference group & Density-based  & Final scoring  \\
 SAFARI-FR-CC &  Mean-Std & Fixed reference group & Clustering-based & Final scoring \\
  SAFARI-FR-FREQ &  SAX & Fixed reference group & Frequency-based & Final scoring \\
    SAFARI-SW-NN & Mean-Std & Sliding Window &  Nearest neighbors-based & Final scoring  
  \\ SAFARI-SW-DEN  & Mean-Std & Sliding Window & Density-based  & Final scoring  \\
 SAFARI-SW-CC &  Mean-Std & Sliding Window & Clustering-based & Final scoring \\
  SAFARI-SW-FREQ &  SAX & Sliding Window & Frequency-based & Final scoring \\
      SAFARI-LW-NN & Mean-Std & Landmark Window &  Nearest neighbors-based & Final scoring  
  \\ SAFARI-LW-DEN  & Mean-Std & Landmark Window & Density-based  & Final scoring  \\
 SAFARI-LW-CC &  Mean-Std & Landmark Window & Clustering-based & Final scoring \\
  SAFARI-LW-FREQ &  SAX & Landmark Window & Frequency-based & Final scoring \\
        SAFARI-URES-NN & Mean-Std & Uniform Reservoir  &  Nearest neighbors-based & Final scoring  
  \\ SAFARI-URES-DEN  & Mean-Std & Uniform Reservoir  & Density-based  & Final scoring  \\
 SAFARI-URES-CC &  Mean-Std & Uniform Reservoir  & Clustering-based & Final scoring \\
  SAFARI-URES-FREQ &  SAX & Uniform Reservoir & Frequency-based & Final scoring \\
          SAFARI-ARES-NN & Mean-Std & Anomaly-aware Reservoir  &  Nearest neighbors-based & Final scoring  
  \\ SAFARI-ARES-DEN  & Mean-Std & Anomaly-aware Reservoir  & Density-based  & Final scoring  \\
 SAFARI-ARES-CC &  Mean-Std & Anomaly-aware Reservoir  & Clustering-based & Final scoring \\
  SAFARI-ARES-FREQ &  SAX & Anomaly-aware Reservoir & Frequency-based & Final scoring \\

 \hline
\end{tabular}
}
\caption{The list of 20 SAFARI detectors as the combination of different SAFARI methods}
\label{detectors}
\end{table}

\section*{Appendix B}
\setcounter{figure}{0}  
\renewcommand{\thefigure}{B\arabic{figure}}
%\captionsetup{labelformat=AppendixFigures}
%\setcounter{figure}{0}

\begin{figure}[H]
    \centering
    \begin{subfigure}[t]{0.5\textwidth}
        \centering
        \includegraphics[height=2.6in]{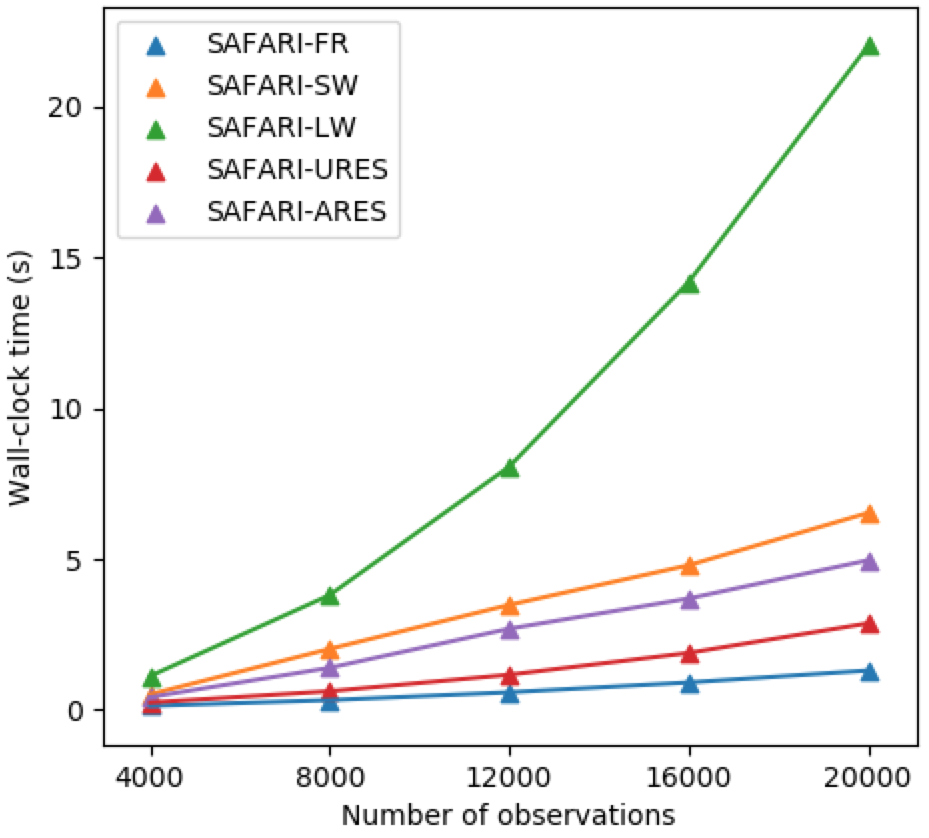}
        \caption{Scalability of learning strategies}
        \label{scalability}
    \end{subfigure}%
    ~ 
    \begin{subfigure}[t]{0.5\textwidth}
        \centering
        \includegraphics[height=2.6in]{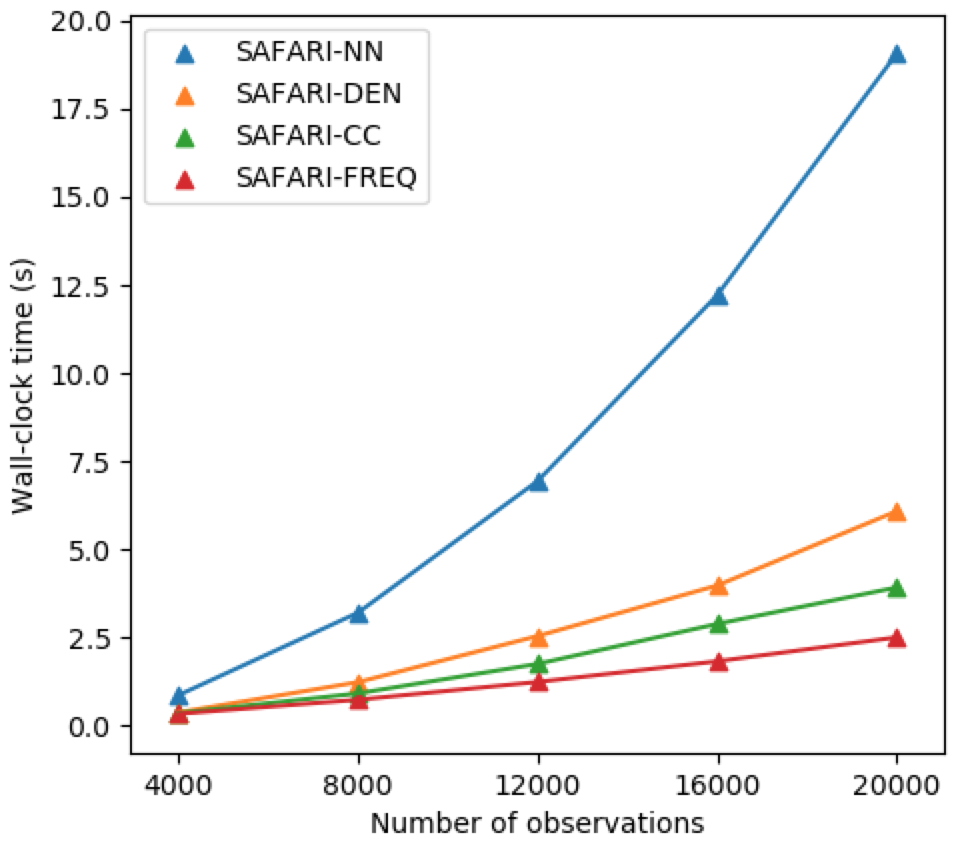}
        \caption{Scalability of nonconformity measures}
        \label{scalability2}
    \end{subfigure}
    \caption{Scalability of different SAFARI methods: wall-clock time of (a) learning strategies (b) nonconformity measure against number of observations in a dataset.}
    \label{sc}
\end{figure}

We empirically studied how runtime of SAFARI methods varies with the dataset size. All experiments were performed on an OSX personal computer with 16GB memory. Runtimes were averaged over five trials. To measure scalability with respect to the size of the dataset, we sampled the number of observations between [4000, 20000] in equal intervals from a single large data file in NAB, consisting of 22,695 data records. The scalability of different learning strategies and nonconformity measures can be seen in Fig. \ref{scalability} and Fig. \ref{scalability2}, respectively.

\begin{figure}[H]
  \centering
    \includegraphics[width=0.8\textwidth]{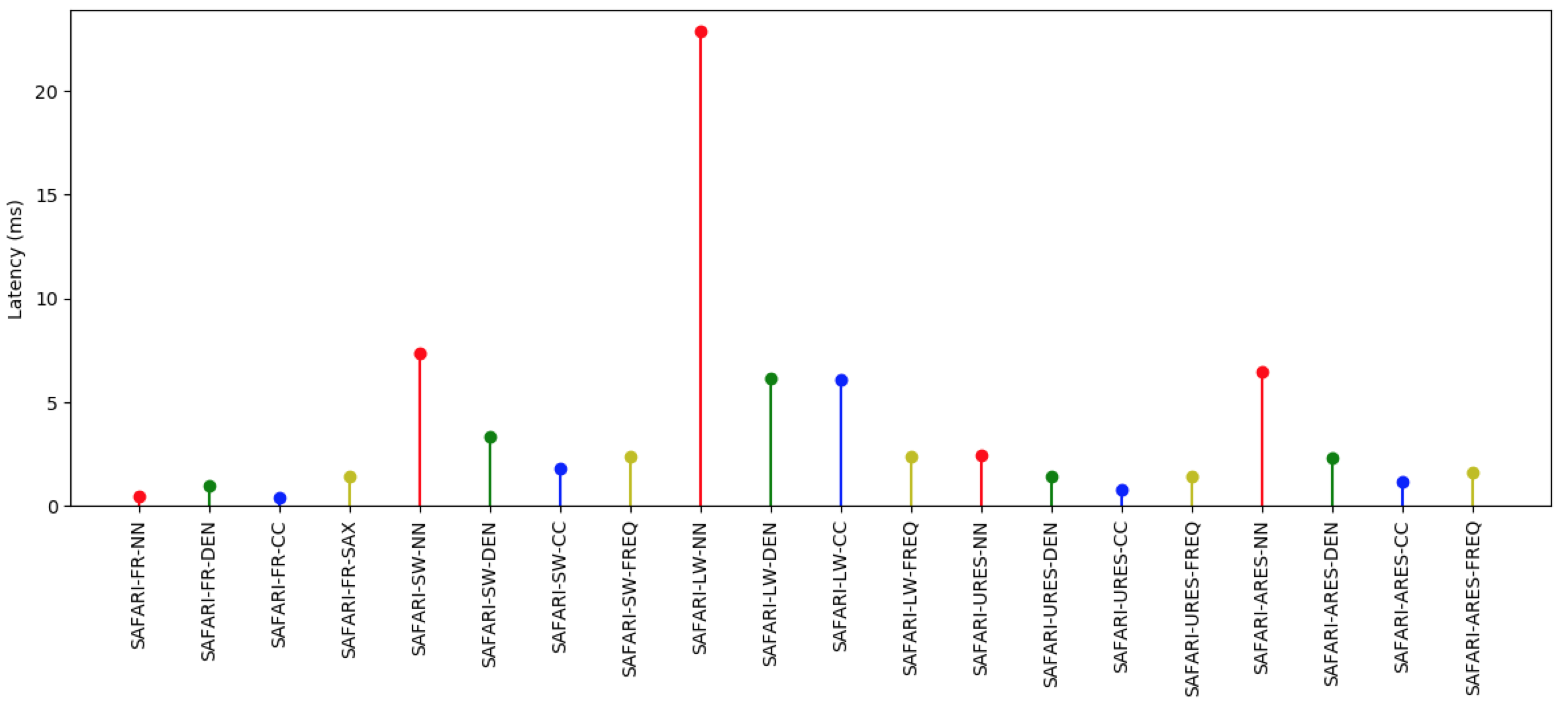}
    \caption{Comparison of 20 SAFARI detectors based on latency (ms).}
     \label{latency}
\end{figure}

Furthermore, we analyzed latency times of each SAFARI detectors used in this study. Latency measures the time taken to process a single data point for anomaly detection. Latency times of detectors are also averaged over 5 runs on the same NAB dataset, consisting of 22,695 data records and shown in Fig. \ref{latency}.

\bibliography{mybib}

\end{document}